\documentclass[12pt]{article}
\usepackage{amsmath}
\usepackage{amsthm}
\usepackage{amssymb}
\usepackage{geometry}
\usepackage{tikz}
\usepackage{algorithm}
\usepackage{algpseudocode}
\usepackage{xcolor}           
\usepackage{tipa}

\usepackage{draftwatermark}
\SetWatermarkText{}
\SetWatermarkScale{2}

\definecolor{LinkBlue}{RGB}{0,114,178}   
\definecolor{CiteOrange}{RGB}{213,94,0}  
\definecolor{URLGreen}{RGB}{0,158,115}   

\usepackage[
  colorlinks = true,
  linkcolor  = LinkBlue,    
  citecolor  = CiteOrange,  
  urlcolor   = URLGreen,    
  filecolor  = URLGreen     
]{hyperref}
\usepackage{doi}                   
\usepackage[
  backend=biber,
  style=numeric,      
  sorting=none,       
  doi=true,
  url=true
]{biblatex}
\usepackage{tcolorbox}
\usepackage{booktabs}  
\usepackage{tabularx}
\usepackage{extarrows} 
\usepackage{subcaption}

\usetikzlibrary{shapes.geometric, arrows.meta, positioning}
\usepackage{chngcntr}
\counterwithout{table}{section}

\theoremstyle{definition}
\newtheorem{definition}{Definition}[section]
\newtheorem{theorem}{Theorem}[section]
\newtheorem{proposition}{Proposition}[section]
\newtheorem{lemma}{Lemma}[section]
\newtheorem{remark}{Remark}[section]
\newtheorem{corollary}{Corollary}[theorem]

\title{A Mathematical Theory of Discursive Networks}
\author{
    Juan B. Gutiérrez\footnote{\copyright 2026 Juan B. Gutiérrez. \href{mailto:juan.gutierrez3@utsa.edu}{juan.gutierrez3@utsa.edu} Department of Mathematics, University of Texas at San Antonio.}  \phantom{ } 
}
\date{\today}

\begin{document}

\maketitle

\begin{abstract}
Large language models (LLMs) turn writing into a live exchange between humans and software. We characterize this new medium as a  discursive network that treats people and LLMs as equal nodes and tracks how their statements circulate. We define the generation of erroneous information as invalidation (any factual, logical, or structural breach) and show it follows four hazards: drift from truth, self-repair, fresh fabrication, and external detection. We develop a general mathematical model of discursive networks that shows that a network governed only by drift and self-repair stabilizes at a modest error rate. Giving each false claim even a small chance of peer review shifts the system to a truth-dominant state.   We operationalize peer review with the open-source \emph{Flaws-of-Others (FOO) algorithm}: a configurable loop in which any set of agents critique one another while a harmonizer merges their verdicts. We identify an ethical transgression, epithesis, that occurs when humans fail to engage in the discursive network. The takeaway is practical and cultural: reliability in this new medium comes not from perfecting single models but from connecting imperfect ones into networks that enforce mutual accountability.
\end{abstract}

\tableofcontents

\section{Introduction}
Large Language Models (LLMs) are advanced artificial intelligence systems trained on vast amounts of text data to generate human-like language. These models utilize deep learning techniques to understand and produce text based on the patterns and structures found in their training data \parencite{vaswani2017attention, brown2020language}. LLMs have demonstrated impressive capabilities in various natural language processing tasks, including text generation, translation, and question-answering.

Despite their remarkable performance, LLMs are prone to generating false or misleading statements, a phenomenon often referred to as ``hallucinations" \parencite{ji2022survey}. However,  the metaphor of hallucination is limited: it implies a private sensory distortion, whereas an unfounded LLM assertion can be circulated, cited, and acted upon as fact.  Throughout this paper we therefore use the broader term
\textbf{invalidation}, and we show that what is commonly referred to as  hallucination is just one of the many manifestations of invalid information.

Invalidations in LLMs can manifest as factual inconsistencies, logical contradictions, or entirely fabricated content that appears plausible \parencite{maynez2020faithfulness}. This issue is exacerbated by the lack of a verification mechanism within the models themselves \parencite{bender2021dangers, weidinger2021ethical}. Although retrieval-augmented generation and self-consistency checks \parencite{lewis2020retrieval,wang2023self} reduce the problem, a substantial share of outputs remains unreliable-enough to undermine trust in practical deployments.

Empirical evaluations show that LLMs continue to produce non-trivial rates of factual error and harmful content after instruction tuning and reinforcement learning. Studies in medical domains have documented significant factual inaccuracies in model answers \cite{thirunavukarasu2023large,sallam2023chatgpt}, and work on adversarial prompting has demonstrated that safety-trained models still emit disallowed content \cite{zou2023universal,chao2023jailbreaking}. 

Several factors contribute to the occurrence of invalid information, including biases in training data, limitations in knowledge representation, and the models' tendency to prioritize fluency over factual accuracy \parencite{lin2022truthfulqa}. The prevalence and impact of invalidations are significant, with quantitative evaluations revealing that they occur in up to 30\% of generated responses, substantially affecting the trustworthiness of these models \parencite{ji2022survey, lin2022truthfulqa}. Moreover, studies have shown that LLMs can generate false information even when explicitly prompted to be truthful \parencite{evans2021truthful}, underscoring the challenge of aligning model outputs with factual correctness.

This article is organized as follows: Section \ref{sec:invalidity-universal} introduces the concept of invalidation as a broader alternative to hallucination; Section \ref{sec:invalidity-human} situates invalidation within established cognitive and media theories, showing it as a universal feature of both human and artificial cognition; Section \ref{sec:spillover-discursive-networks} examines how invalidations propagate through interconnected communication systems; Section \ref{sec:efficient-engagement} provides an information-theoretic basis for why verification is fundamentally easier than generation; Section \ref{sec:discursive} formalizes discursive networks as mathematical structures with actors, statements, and update rules; Section \ref{sec:modeling} develops three progressively complex models of invalidation dynamics: single-network with binary states, single-network with emergent invalidation, and cross-network detection; Section \ref{sec:theoretical-validation} demonstrates the mathematical consistency of these models through theoretical analysis and parameter exploration; finally, Section \ref{sec:conclusion} addresses ethical concerns including epithesis, energy costs, and epistemic diversity, before outlining future research directions.

The scope of this manuscript encompasses both theoretical foundations and practical implementations. We establish mathematical floors on invalidation probability, develop network models for error propagation and detection, and present the Flaws-of-Others (FOO) algorithm with cryptographic integrity verification. The analysis focuses on invalidations that emerge during inference in large language models and mitigation strategies based on cross-agent critique. While we do not provide exhaustive failure catalogues or benchmark comparisons, we advance a unified mathematical framework demonstrating how networks of imperfect agents can achieve error rates below what any individual agent attains.

\subsection{From ``Hallucination" to Invalidation}
\label{sec:invalidity-universal}

The word \emph{hallucination} has become the default label for false
statements generated by LLMs.  Borrowed from perceptual psychology, it
misses two crucial aspects.  First, an LLM's error is not confined to a
private experience; it can be adopted by readers and propagate through
networks, amplifying misinformation \parencite{crawford2021excavating}.
Second, focussing on hallucinations alone narrows the research agenda,
leaving other error classes (logical contradictions, format violations,
ethical breaches) under-examined.

We call any output that violates a constraint set of facts, logic, norms, and formats,  
an \emph{invalidation}.  Because an autoregressive decoder maximises
next-token likelihood rather than global consistency, a non-zero slice
of probability mass inevitably falls outside the constraints. 

In practice, \emph{invalidation} surfaces along at least five recurring archetypes that differ in locus and detectability.  
First, \textbf{hallucination} denotes the introduction of content that is ungrounded in any trusted source or context; large-scale surveys show it to be pervasive even in the highest-performing models \parencite{huang2025survey}.  
Second, \textbf{contradiction} captures internally inconsistent statements that coexist within a single generation, a failure mode quantified and mitigated by prompt-based self-refinement techniques \parencite{mundler2023selfcontradiction}.  
Third, \textbf{deductive error} arises when the model draws logically invalid conclusions from true premises, an error family systematically stress-tested with adversarial perturbations \parencite{hoppe2025deductive}.  
Fourth, \textbf{pragmatic impropriety} concerns outputs that violate social or professional norms, including toxicity, hate speech, or privacy leakage; the \emph{RealToxicityPrompts} benchmark revealed that even innocuous inputs can trigger toxic degeneration \parencite{gehman2020realtoxicity}.  
Finally, \textbf{format violation} occurs when the model breaks explicit structural constraints (e.g., JSON Schema), jeopardising downstream machine consumption; work with \emph{JSONSchemaBench} shows that such violations remain stubbornly frequent despite constrained decoding \parencite{geng2025jsonschema}.

Each error class manifests one predicate: the content fails to match a state of the world. That predicate admits representation with an given invalidation rate. The models in Section \ref{sec:discursive} use the rate alone and does not depend on class labels while it quantifies the chance that any output lacks validity. 

Taken together, these archetypes point to a broad invalidation family, potentially with more members,  underscoring the need for evaluation suites and mitigation strategies that address the full spectrum of failure modes. Figure~\ref{fig:taxonomy} schematizes this superset-subset relationship. 

\usetikzlibrary{positioning}

\begin{figure}[t]
  \centering
  \begin{tikzpicture}[scale=0.9]

    \draw[line width=1pt] (0,0) circle (4cm);
    \node[font=\large\bfseries] at (0,4.55) {Invalid Outputs};

    \begin{scope}[every node/.style={font=\footnotesize,align=center}] 
      \filldraw[fill=blue!20,draw=black] (-1.6,1.9) circle (1.2cm);
      \node at (-1.6,1.9) {Hallucination};

      \filldraw[fill=green!20,draw=black] (1.6,1.9) circle (1.2cm);
      \node at (1.6,1.9) {Contradiction};

      \filldraw[fill=orange!25,draw=black] (0,-0.2) circle (1.2cm);
      \node at (0,-0.2) {Deductive\\Error};

      \filldraw[fill=purple!20,draw=black] (-2.1,-1.6) circle (1.2cm);
      \node at (-2.1,-1.6) {Pragmatic\\Impropriety};

      \filldraw[fill=red!25,draw=black] (2.2,-1.5) circle (1.2cm);
      \node at (2.2,-1.5) {Format\\Violation};
    \end{scope}

  \end{tikzpicture}
  \caption{Venn-style illustration of members of a broader set of invalid outputs produced by large language models.}
  \label{fig:taxonomy}
\end{figure}

\subsection{Categorization of Mechanisms to Engage LLM Agents}
\label{sec:efficient-engagement}
We now examine how different engagement mechanisms map onto the information-theoretic landscape established above. During inference, a Transformer language model executes a single forward pass per token. The model transforms context $\mathbf{x}_{\le t}$ into embeddings, applies $L$ fixed self-attention layers to produce hidden state $\mathbf{h}_t$, and maps this to a token probability distribution via:
\[
p(w_{t+1}\mid\mathbf{x}_{\le t}) = \mathrm{softmax} \bigl(W\mathbf{h}_t\bigr).
\]
No gradient updates or objective-function evaluations occur at this stage; the only on-the-fly ``optimization'' is the decoding heuristic (greedy, top-$k$, nucleus, or beam search) that selects the next token from the static distribution $p(\cdot)$.

LLMs support a range of functional output types that can be systematically grouped into three core categories we introduce in this manuscript: \emph{ideation} (constructive synthesis), \emph{critique} (diagnostic evaluation), and \emph{product-oriented generation} (goal-directed deliverables). These three umbrella functions organize diverse tasks, each with distinct inference properties, compositional demands, and distributional positions in the model's output space. Existing literature identifies several subtypes that map naturally onto these categories \parencite{bommasani2021opportunities, mialon2023augmented}.

\textbf{Critique output} is structurally efficient to generate. 
Nested within critique are:
\begin{itemize}
    \item \emph{Flaws of others}. This approach, introduced as a mechaism for verification in this mansucript (cfr. Section \ref{sec:foo-algorithm}) favors the strategy to ask for flaws  or errors in the outputs of other agents, resulting in constructions 
    that are frequent in training corpora thus placing them near the peak of $p(\cdot)$. 
    Even shallow decoding heuristics retrieve frequent patterns with high fluency and relevance \parencite{wei2022chain}.
  \item \emph{Classification and Disambiguation}, such as assigning sentiment, stance, or intent. These tasks resolve ambiguity and often underlie evaluation pipelines \parencite{mialon2023augmented}.
  \item \emph{Restatement and Summarization}, which surface structural coherence or hidden biases by rephrasing or compressing content. When used diagnostically, they reveal implicit assumptions or inconsistencies \parencite{maynez2020faithfulness}.
\end{itemize}

\textbf{Ideation output} demands compositional novelty. Prompts that ask the model to hypothesize mechanisms, imagine alternatives, or propose designs typically land in the tail of the output distribution. Generating them requires broader exploration (via large beam width or elevated temperature) and exhibits greater output variance.

Within ideation, we find:
\begin{itemize}
  \item \emph{Instruction and Procedural Guidance}, where the model scaffolds user understanding or explains concepts in sequence. These tasks require didactic clarity and often invoke implicit audience modeling \parencite{ouyang2022training}.
  \item \emph{Meta-Reasoning and Strategy Output}, which includes multi-step planning, evaluating hypotheses, or chain-of-thought reasoning. These outputs require recursive coherence and longer dependency tracking \parencite{wei2022chain}.
\end{itemize}

\textbf{Product-oriented output} targets the generation of external artifacts: source code, formatted markup, structured data, or interactive dialogue. These tasks often carry hard constraints and precision demands. Simple forms (e.g., boilerplate code) reside in high-probability zones, while structurally complex or compositional outputs require deeper exploration.

Included in this class are:
\begin{itemize}
  \item \emph{Formalism Translation}, such as converting text to e.g. JSON, SQL,  \LaTeX, etc. This requires syntax-aligned generation and tight coupling between prompt and output form \parencite{reynolds2021prompt}.
  \item \emph{Retrieval-Simulation}, where the model reproduces facts or references learned during pretraining. These outputs appear fluent but are not grounded in current truth, making them useful but epistemically fragile \parencite{bommasani2021opportunities}.
  \item \emph{Social Interaction Simulation}, which includes emulating customer support, roleplay, or therapeutic dialogue. These are product-like in that the output is consumed as experience or interface, and they require tone, persona, and context alignment \parencite{jo2025proxyllm, park2023generative, song2024typing}.
\end{itemize}



\subsection{Invalidation as a Universal Feature of Human and Artificial Cognition}
\label{sec:invalidity-human}


Human cognition systematically prioritizes narrative coherence over factual accuracy, a tendency that emerges not from individual pathology but from the fundamental architecture of meaning-making itself. When confronted with contradictory evidence, both individuals and groups construct elaborate justifications that preserve existing belief structures rather than revising them \parencite{festinger1957cognitive, cohen2001states}. This preference for coherence manifests across scales: from the micro-level impression management that shapes everyday social interactions \parencite{goffman1959presentation} to the macro-level collective narratives that enable societies to ignore systemic atrocities \parencite{cohen2001states}. Remarkably, this same structural bias toward local coherence over global accuracy appears in large language models, where autoregressive architectures favor maintaining consistency with previous tokens even at the expense of factual correctness. The parallel suggests that invalidation arises not from a flaw in either human or artificial systems, but from a deeper computational trade-off inherent to any agent that must construct meaning from sequential, uncertain information.

The medium itself acts as an epistemic filter, determining not just what information reaches us but what we accept as real, a process that operates identically whether the medium is television, print journalism, or a large language model. Media theorists have long recognized that truth emerges less from content evaluation than from structural repetition: the same claim, encountered repeatedly through trusted channels, eventually sediments into accepted fact regardless of its veracity \parencite{gerbner1976living, mcluhan1964understanding}. This manufacturing of consensus operates through cascading filters (economic incentives, institutional biases, and technological affordances) that systematically amplify certain narratives while suppressing others \parencite{herman1988manufacturing}. Large language models instantiate this same filtering mechanism at an unprecedented scale: their training corpora encode the biases of millions of sources, their attention mechanisms privilege frequently repeated patterns, and their optimization objectives reward fluent reproduction over factual verification. The result is a computational echo chamber where invalidations, once embedded in training data, achieve the same truth-like status through sheer statistical dominance that media repetition grants to human beliefs.

Propaganda operates by exploiting a fundamental vulnerability in epistemic systems: sustained repetition of coordinated falsehoods eventually overwhelms the capacity for empirical verification, creating an alternate reality that becomes self-reinforcing through social proof. This mechanism, which political theorists identify as the cornerstone of totalitarian control, functions by flooding the information environment with internally consistent but externally false narratives until the sheer cognitive cost of maintaining skepticism exceeds most people's capacity \parencite{arendt1951origins, ellul1965propaganda}. The parallel with large language models is striking: trained on billions of documents where certain false narratives appear thousands of times, these systems internalize misinformation not through ideological commitment but through pure statistical frequency. Just as propaganda succeeds by making lies more cognitively available than truth \parencite{ellul1965propaganda}, LLMs generate invalidations by sampling from probability distributions where well-represented falsehoods outweigh poorly-documented facts. The computational architecture thus recreates, without intention or awareness, the same reality-distortion mechanisms that human propagandists deploy deliberately.

Invalidation emerges from the fundamental computational shortcuts that make complex reasoning tractable; these shortcuts manifest identically in biological neural networks and artificial transformers. The core mechanism is substitution: when faced with difficult questions about truth or probability, both humans and LLMs unconsciously replace them with easier questions about familiarity and similarity \parencite{tversky1974judgment, kunda1990motivated}. This substitution operates through dual channels: the availability heuristic replaces ``what is true?" with ``what comes easily to mind?", while the representativeness heuristic replaces ``what is probable?" with ``what resembles my prototype?" In large language models, these exact substitutions occur mechanistically, e.g. the softmax function literally converts truth-seeking into frequency-matching, while attention heads select tokens based on similarity rather than veracity. The result is a convergent failure mode where both human reasoning and machine generation systematically mistake statistical patterns for factual reality, producing confident invalidations that feel true precisely because they align with existing distributions rather than external facts \parencite{kunda1990motivated}.

The cross-disciplinary convergence of these observations suggests that invalidation 
%
%
is not simply a by-product of autoregressive sampling, but a potentially universal cognitive process rooted in the fundamental nature of information processing through language. 
%
%
%
%
Rather than indicating a breakdown of alignment, it may reveal the presence of alignment to socially and contextually shaped heuristics that guide behavior in uncertain conditions, bounded rationality, and social constraints.

Addressing invalidation in LLMs will require approaches that incorporate sociological, psychological, and media-theoretical models of belief formation and narrative control. At the same time, observing how LLMs generate invalidations may provide new empirical traction for understanding human cognitive phenomena such as confirmation bias, belief perseverance, and collective denial. The interdependence between artificial and human cognition in this respect suggests that solutions to the propagation of invalidation may emerge not solely from engineering but from a broader inquiry into the structure of meaning-making itself.

\subsection{Risks of Invalidation Spill-over Through Discursive Networks}
\label{sec:spillover-discursive-networks}
Invalidations produced by large language models might not remain isolated.
Once released into public channels they could propagate through
interconnected systems of communication.  A
\emph{discursive network} is a large-scale ecosystem of human and
machine agents whose utterances circulate, reinforce, and mutate through
repeated exchange.  When an LLM instantiates many synthetic voices that
emit high volumes of text, those voices become additional nodes that
mediate narrative transmission and amplification.

We use the term \emph{discursive network}, rather than the established \emph{discourse network}, to signal a McLuhanian twist.  In classical discourse-network analysis the medium is a passive conduit: speeches, papers, and news items are shuffled among human actors while the underlying carrier remains inert.  In the networks that include LLMs the carrier intervenes.  The medium does not just move the message; it edits, rewrites, and recombines it at every hop.  By switching from \emph{discourse} to \emph{discursive} we emphasise that language itself is now produced inside the network dynamics, co-authored by the very infrastructure that transmits it; this is a direct extension of McLuhan's dictum from ``the medium is the message'' to ``the medium acts on the message.''

Discursive networks consist of nodes (agents or messages) and edges
(interactions, citations, reshares).  
Closely related structures appear in political-science
\emph{discourse-network analysis}, where policy actors and speech acts
form time-evolving coalitions \parencite{leifeld2014,leifeld2012}.
Kittler's media archaeology likewise foregrounds how technological
substrates shape what counts as meaningful speech
\parencite{kittler1990}.  
Our focus on generative models extends these traditions by treating LLM
outputs as first-class network nodes.

Humans struggle to separate machine-generated prose from human writing.
Controlled studies across genres report identification accuracy only
slightly above chance (about 55-65 \%) and show that readers rely on
fragile surface heuristics such as pronoun frequency and stylistic
fluency \parencite{ippolito2020,DBLP:conf/acl/GehrmannSR19}.  
In a discursive network this limitation matters: synthetic
invalidations, mis-taken for human statements, are more readily reposted
or cited.

Once injected, invalid content spreads via echo-amplification.
Research on social-media communities finds that clusters organized in
bow-tie topologies (tightly knit cores with radiating peripheries) amplify
low-quality or misleading messages more than high-quality ones
\parencite{garimella2018}.  Agents in these clusters can absorb and relay
invalidations, algorithmic or human, without verification, entrenching
flawed narratives.

Detection tools provide only partial relief.  State-of-the-art AI
detectors fall below 80 \% accuracy on paraphrased or adversarial
samples and show biases toward false positives on human text
\parencite{huang2024robust,tufts2024practical,sadasivan2025reliably}.  Human moderators fare little better, typically
around 65 \% in blinded settings \parencite{ippolito2020}.  Because both
machines and people misclassify a substantial share of content,
invalidations re-enter discourse with minimal friction.

Mitigating spill-over therefore demands a synthesis of discourse-network
analysis, community-structure research, and detection studies.  High-
volume clusters of unverified claims need tracing, and interventions
should target influential nodes, always within robust privacy and
governance constraints.  Recognising synthetic agents as network
participants reframes classic media theory for the generative era:
discursive-network analysis becomes both a descriptive lens on
spill-over risk and a practical framework for targeted intervention.

\section{Methods}
This section lays out the analytical and computational machinery that supports
our theory of discursive networks.  We begin with a distribution-relative lower
bound on invalidation probability: if a reference distribution assigns positive
mass to invalid outputs and a model remains at finite Kullback--Leibler
divergence from that reference, then the model must assign positive probability
to that invalid set.  This bound is a support-and-divergence statement.  It does
not preclude deployment layers---retrieval, constrained decoding, rejection
sampling, formal verification, or human review---from changing the delivered
distribution.

The lower bound motivates verification layers, but verification advantage is not
a universal entropy theorem.  We therefore model verification as a noisy channel
with measurable sensitivity, false-positive rate, repair probability,
dependence, and common-mode blind spots.  Under this model, discursive networks
reduce delivered invalidity when critics detect a nonzero fraction of
non-common-mode errors and flagged errors are successfully repaired or rejected.
They cannot overcome shared blind spots merely by adding more replicas of the
same critic family.

The remainder of the section builds the network model.  We formalize discursive
networks as structures $N=(S,P,I,C,A,B,U,G)$ whose agents exchange and validate
statements under well-defined update rules (Section~\ref{sec:discursive}).  We
then analyze two-state belief dynamics under a single fixed-population framework
(Section~\ref{sec:modeling}), using two effective parameters: $a$ for
truth-to-false pressure and $b$ for false-to-true correction.  The familiar
parameters $p,q,\lambda,d$ enter as special cases of $a$ and $b$.  We distinguish
exact discrete-time Bernoulli detection from continuous-time additive hazards,
because the minimum-agent formula differs in the two regimes.  Finally, the
Flaws-of-Others (FOO) algorithm implements conditional cross-agent verification
and records the interaction history through a tamper-evident signed hash-linked
log.

\subsection{A distribution-relative floor on invalidation probability}
\label{sec:invalidation-floor}

Let $(\Omega,\mathcal F)$ be the measurable space of finite text outputs,
including an end-of-sequence token.  A discursive standard is represented by a
measurable validity predicate
\[
    C:\Omega\to\{0,1\},
\]
where $C(x)=1$ means that $x$ satisfies the relevant standard and $C(x)=0$ means
that $x$ is invalid.  The invalid set is
\[
    S=\{x\in\Omega:C(x)=0\}.
\]
Let $Q$ denote a reference distribution, such as the distribution induced by a
training corpus, task data, or evaluation environment, and let $P_\theta$ denote
the model distribution being evaluated.  To avoid collision with the dynamical
self-repair probability later denoted by $q$, write
\[
    q_{\mathrm{corp}}=Q(S),\qquad
    p_\theta=P_\theta(S),\qquad
    D=D_{\mathrm{KL}}(Q|P_\theta).
\]
The event $S$ is a coarse-graining of the full sequence space.  Applying the
data-processing inequality for KL divergence to the binary map
$x\mapsto\mathbf 1_{\{x\in S\}}$ gives the appropriate lower bound
\parencite{cover2006elements}.

\begin{lemma}[Event lower bound from relative entropy]
\label{lem:invalidation-floor}
Assume $q_{\mathrm{corp}}=Q(S)>0$ and
$D_{\mathrm{KL}}(Q|P_\theta)\le D<\infty$.  Let
\[
 d_{\mathrm{Bern}}(q|p)
 =q\log\frac{q}{p}+(1-q)\log\frac{1-q}{1-p}
\]
denote binary KL divergence, with natural logarithms.  Then
\[
 d_{\mathrm{Bern}}\bigl(q_{\mathrm{corp}}|P_\theta(S)\bigr)\le D.
\]
Consequently $P_\theta(S)>0$.  More sharply, if
$\beta(q_{\mathrm{corp}},D)\in(0,q_{\mathrm{corp}}]$ is the unique solution of
\[
 d_{\mathrm{Bern}}\bigl(q_{\mathrm{corp}}|\beta(q_{\mathrm{corp}},D)\bigr)=D,
\]
then
\[
    P_\theta(S)\ge \beta(q_{\mathrm{corp}},D).
\]
A fully explicit, slightly weaker bound is
\[
 P_\theta(S)
 \ge
 \exp\left(-\frac{D+h_2(q_{\mathrm{corp}})}{q_{\mathrm{corp}}}\right)
 =q_{\mathrm{corp}}(1-q_{\mathrm{corp}})^{(1-q_{\mathrm{corp}})/q_{\mathrm{corp}}}
   \exp(-D/q_{\mathrm{corp}}),
\]
where $h_2(q)=-q\log q-(1-q)\log(1-q)$ is the binary entropy in nats.
\end{lemma}

\begin{proof}
Apply the measurable map $\varphi(x)=\mathbf 1_{\{x\in S\}}$.  The pushforwards
of $Q$ and $P_\theta$ under $\varphi$ are Bernoulli distributions with
parameters $q_{\mathrm{corp}}$ and $p_\theta=P_\theta(S)$, respectively.  The
data-processing inequality gives
\[
D_{\mathrm{KL}}(Q|P_\theta)
\ge
D_{\mathrm{KL}}\bigl(\mathrm{Bern}(q_{\mathrm{corp}})|\mathrm{Bern}(p_\theta)\bigr)
=
d_{\mathrm{Bern}}(q_{\mathrm{corp}}|p_\theta).
\]
Since $q_{\mathrm{corp}}>0$, the binary divergence is infinite when
$p_\theta=0$, so finite KL divergence forces $p_\theta>0$.  For fixed
$q\in(0,1]$, the function $p\mapsto d_{\mathrm{Bern}}(q|p)$ is continuous and
strictly decreasing on $(0,q]$, with limit $+\infty$ as $p\downarrow0$ and
value $0$ at $p=q$.  Hence the inverse lower bound follows.  For the explicit
bound, write
\[
 d_{\mathrm{Bern}}(q|p)
 =-h_2(q)-q\log p-(1-q)\log(1-p)
 \ge -h_2(q)-q\log p,
\]
because $-\log(1-p)\ge0$.  Since $d_{\mathrm{Bern}}(q|p)\le D$, we obtain
$-h_2(q)-q\log p\le D$, and therefore
$p\ge \exp\{-(D+h_2(q))/q\}$.
\end{proof}

\begin{remark}[Absolute continuity]
The finite-divergence assumption $D_{\mathrm{KL}}(Q|P_\theta)<\infty$ implies
$Q\ll P_\theta$: every event with positive $Q$-mass must also have positive
$P_\theta$-mass.  This is the relevant support condition.  In contrast,
$P_\theta\ll Q$ would require the model to place zero probability on every
string absent from the reference distribution, which is usually not appropriate
when $Q$ is a finite empirical corpus and $P_\theta$ is a softmax language model.
\end{remark}

\paragraph{Relation to sharper hallucination-floor results.}
Lemma~\ref{lem:invalidation-floor} is deliberately modest: it is a
support-and-divergence statement.  Stronger lower bounds are available for base
language models under additional training and calibration assumptions.  Kalai
and Vempala show that calibrated language models must hallucinate on certain
arbitrary facts at a rate controlled by the Good--Turing missing-mass or
``monofact'' rate of the training corpus \parencite{kalai2024calibrated}.
Kalai, Nachum, Vempala, and Zhang connect generative errors to an
\emph{Is-It-Valid} (IIV) binary classification problem, showing that validity
classification error induces lower bounds on base-model generation error under
their distributional assumptions \parencite{kalai2025why}.  These results
motivate the architecture below: make validity checking explicit, then model its
reliability through sensitivity, false positives, repair success, and
common-mode blind spots.

\paragraph{Interpretation.}
If a reference distribution $Q$ contains invalid outputs with
$q_{\mathrm{corp}}>0$, and a model remains close to $Q$ in KL divergence, then
the model cannot remove all probability mass from the invalid set.  The theorem
is not an absolute floor for every possible deployment pipeline.  A system can
lower delivered error by changing the sampling distribution, retrieving external
evidence, rejecting candidates, enforcing formal constraints, or routing outputs
through verifiers.  The mathematical question therefore shifts from whether one
model can be perfect to the conditions under which verifier networks reduce
residual invalidity.

\subsection{Verification advantage as a conditional channel property}
\label{sec:verification-advantage}

Verification is not universally easier than generation.  Some false claims are
easy to detect because they violate syntax, arithmetic, cited evidence, a formal
specification, or well-indexed facts.  Other false claims require essentially
the same expertise as generating the correct answer.  Modern agentic systems
exploit verification most successfully when they add external evidence, tools,
learned outcome verifiers, process reward models, debate, self-refinement, or
formal checks
\parencite{cobbe2021trainingverifiers,uesato2022processoutcome,lightman2023verify,irving2018debate,du2023multiagentdebate,yao2022react,schick2023toolformer}.  Conversely,
intrinsic self-correction without external feedback can fail or degrade
performance, and verifier effectiveness varies with problem difficulty,
generator capability, and verifier capability
\parencite{huang2023cannotselfcorrect,zhou2025verificationdynamics}.

The chain-rule identity for variable-length outputs is useful as bookkeeping,
but it is not by itself a theorem of verification advantage.

\begin{lemma}[Variable-length entropy decomposition]
\label{lem:variable_length_decomposition}
Let $(L_g,Y_g)$ and $(L_v,Y_v)$ be the joint length--content random variables
for generation and verification tasks, respectively.  Let
$\pi^{\mathrm{len}}_g(n)=\Pr(L_g=n)$ and
$\pi^{\mathrm{len}}_v(n)=\Pr(L_v=n)$.  Then, with all logarithms in the same
base,
\begin{align*}
H(L_g,Y_g)-H(L_v,Y_v)
&=H(L_g)-H(L_v) \\
&\quad +\sum_n \pi^{\mathrm{len}}_g(n)H(Y_g\mid L_g=n)
       -\sum_n \pi^{\mathrm{len}}_v(n)H(Y_v\mid L_v=n).
\end{align*}
\end{lemma}

\begin{proof}
Apply $H(X,Y)=H(X)+H(Y\mid X)$ to $(L_g,Y_g)$ and $(L_v,Y_v)$ and subtract.
\end{proof}

\begin{definition}[Verification channel]
\label{def:verification-channel}
For a generated claim, let $E\in\{0,1\}$ indicate invalidity, with
\[
 \alpha=\Pr(E=1).
\]
Let $U\in\{0,1\}$ indicate a common-mode blind spot for the critic family under
consideration.  Conditional on invalidity,
\[
  c=\Pr(U=1\mid E=1).
\]
If $U=1$, this critic family has no signal for the error.  For critic
$i=1,\dots,m$, let $F_i\in\{0,1\}$ indicate that critic $i$ flags the claim.
For non-common-mode invalid claims, define the aggregate detection probability
\[
  \delta_m=\Pr\left(\bigcup_{i=1}^m\{F_i=1\}\,\middle|\,E=1,U=0\right).
\]
For valid claims, define the aggregate false-positive probability
\[
  \eta_m=\Pr\left(\bigcup_{i=1}^m\{F_i=1\}\,\middle|\,E=0\right).
\]
When the critics are conditionally independent, with sensitivities
$s_i=\Pr(F_i=1\mid E=1,U=0)$ and false-positive rates
$r_i=\Pr(F_i=1\mid E=0)$, these reduce to
\[
  \delta_m=1-\prod_{i=1}^m(1-s_i),
  \qquad
  \eta_m=1-\prod_{i=1}^m(1-r_i).
\]
\end{definition}

\begin{theorem}[Residual invalidity after cross-verification]
\label{thm:verification-channel}
Under Definition~\ref{def:verification-channel}, suppose a flagged invalid claim
is successfully repaired or rejected with probability $\rho\in[0,1]$, and the
repair procedure does not introduce new invalid claims.  Then the residual
invalidity rate per generated claim after one verification round is
\[
\alpha_{\mathrm{out}}
=
\alpha\bigl[c+(1-c)(1-\rho\delta_m)\bigr].
\]
The expected false-positive review burden per generated claim is
\[
\phi_{\mathrm{fp}}=(1-\alpha)\eta_m,
\]
and the precision of a flag is
\[
\Pr(E=1\mid \exists i:F_i=1)
=
\frac{\alpha(1-c)\delta_m}
{\alpha(1-c)\delta_m+(1-\alpha)\eta_m}.
\]
Under conditional independence, substitute
$\delta_m=1-\prod_i(1-s_i)$ and $\eta_m=1-\prod_i(1-r_i)$.
\end{theorem}

\begin{proof}
An invalid claim in the common-mode blind spot contributes residual mass
$\alpha c$.  For an invalid claim outside that blind spot, the verification
ensemble flags it with probability $\delta_m$, and a flag removes or repairs the
invalidity with probability $\rho$.  The residual non-common-mode contribution
is therefore $\alpha(1-c)(1-\rho\delta_m)$.  Adding the two invalid cases gives
$\alpha_{\mathrm{out}}$.  The false-positive burden follows by applying the
same flag event under $E=0$.  The precision formula is Bayes' rule.
\end{proof}

\begin{corollary}[Homogeneous independent critics]
\label{cor:homogeneous-critics}
If $s_i=s$, $r_i=r$, the critics are conditionally independent, and $\rho=1$,
then
\[
\alpha_{\mathrm{out}}(m)=\alpha\bigl[c+(1-c)(1-s)^m\bigr].
\]
A target $\alpha_{\mathrm{out}}(m)\le\varepsilon$ is attainable by this critic
family only if $\varepsilon>\alpha c$.  When
$\alpha c<\varepsilon<\alpha$,
\[
 m_{\min}
 =
 \left\lceil
 \frac{\log\left((\varepsilon/\alpha-c)/(1-c)\right)}{\log(1-s)}
 \right\rceil.
\]
If $\varepsilon\le\alpha c$, adding more critics from the same blind-spot family
cannot reach the target; one must reduce $c$ by adding genuinely new evidence
sources, tools, modalities, human expertise, independent training data, or
formal checks.
\end{corollary}

\paragraph{Decision-theoretic interpretation.}
Classical binary decision theory explains why validity checking is the right
primitive to isolate.  In a two-hypothesis problem with valid-candidate
distribution $P_+$ and invalid-candidate distribution $P_-$, Le Cam's method
bounds the Bayes risk of any binary verifier in terms of the total variation
distance between $P_+$ and $P_-$ \parencite{lecam1973,tsybakov2009}.  This is a
lower bound on binary testing difficulty, not a proof that generation is always
harder.  The IIV reduction of Kalai, Nachum, Vempala, and Zhang gives the modern
language-model analogue: validity-classification error induces lower bounds on
base-model generative error under their distributional assumptions
\parencite{kalai2025why}.  The channel parameters $(\delta_m,\eta_m,c,\rho)$,
or their independent-critic summaries $(s_i,r_i,c,\rho)$, are the operational
quantities that must be estimated in any deployed FOO system.

\paragraph{Implications for discursive networks.}
The theorem gives a conditional verification-advantage claim.  Cross-agent
critique reduces delivered invalidity when sensitivities are positive, repair
probability is nontrivial, false-positive burden is manageable, and common-mode
blind spots are small.  It does not promise arbitrary error reduction when
agents share the same training data, missing evidence, prompt framing, or
reasoning pathology.  This formulation identifies exactly which quantities must
be measured or engineered.

\subsection{Discursive Network Formalization}\label{sec:discursive}
To systematically study the phenomenon of invalidation, we propose a formal model that quantifies how invalidations propagate within a discursive network. This model considers actors as nodes in a network, with edges representing the exchange of statements. The goal is to understand how actors influence each other and how invalidations spread, and if and how it reaches an equilibrium state. The subsequent sections will detail the analytical machinery, yielding quantiative information, for studying invalidation in both human and artificial contexts.

\begin{definition}\label{def:discourse}
    \textbf{Discourse.} In the context of a discursive network, discourse refers to the structured process of communication and interaction between actors, $A = \{a_1, a_2, \ldots, a_n\}$, through which they exchange, validate, invalidate, and attempt to persuade each other regarding the truth or falsity of a set of statements $S = \{s_1, s_2, \ldots, s_m\}$. Discourse encompasses all forms of communication $C = \{C_{ij}\}$ between actors, where beliefs $B_i \subseteq S$ are shared, challenged, or reinforced, as well as the mechanisms of invalidation $I = \{I_{ij}\}$, and persuasion $P = \{P_{ij}\}$, which influence the evolution of each actor's belief set. The outcome of discourse is governed by the update rules $U_j$, which dictate how actors revise their beliefs based on the interactions they engage in.
    \end{definition}

\begin{definition}\label{def:discursive_network}
    \textbf{Discursive Network.} A discursive network is a formal structure $N = (S,P,I,C,A,B,U,G)$ where:
    \begin{itemize}
        \item $S = \{s_1, s_2, \ldots, s_m\}$ is the set of possible statements, where each statement can be either true or false.
        \item $P = \{P_{ij} \mid i, j \in \{1, 2, \ldots, n\}\}$ represents the persuasion functions, where $P_{ij}(s_k)$ gives the likelihood that actor $a_j$ will adopt statement $s_k$ after receiving communication from actor $a_i$.
        \item $I = \{I_{ij} \mid i, j \in \{1, 2, \ldots, n\}\}$ denotes invalidations, where $I_{ij}(s_k, s_l)$ signifies actor $a_i$ invalidating a statement $s_k$ held by actor $a_j$ using a contradictory statement $s_l$.
        \item $C = \{C_{ij} \mid i, j \in \{1, 2, \ldots, n\}\}$ represents the communications between actors, where $C_{ij}$ is the set of statements communicated from actor $a_i$ to actor $a_j$.
        \item $A = \{a_1, a_2, \ldots, a_n\}$ is the set of actors participating in the discourse.
        \item $B = \{B_1, B_2, \ldots, B_n\}$ represents the belief sets of the actors, where $B_i \subseteq S$ denotes the set of statements believed to be true by actor $a_i$.
        \item $U = \{U_j \mid j \in \{1, 2, \ldots, n\}\}$ is the set of update rules that define how each actor's belief set $B_j$ is modified in response to communications and invalidations.
        \item $G = \{G_1, G_2, \ldots, G_n\}$ represents the goal functions of the actors, with $G_i: A \to 2^S$ specifying the set of statements actor $a_i$ seeks to convince other actors to believe.
    \end{itemize}
    The discursive network models the dynamics of belief formation, communication, persuasion, and invalidation among actors within a formal discourse setting.
\end{definition}


\textbf{Example.} Consider a simple scenario with three actors $A = \{a_1, a_2, a_3\}$ and two statements $S = \{s_1, s_2\}$. Actor $a_1$ believes $s_1$ ($B_1 = \{s_1\}$) and wants $a_2$ and $a_3$ to also believe $s_1$ ($G_1(a_2) = G_1(a_3) = \{s_1\}$). Actor $a_2$ believes $s_2$ ($B_2 = \{s_2\}$) and wants $a_1$ and $a_3$ to believe $s_2$ ($G_2(a_1) = G_2(a_3) = \{s_2\}$). Actor $a_3$ is initially neutral ($B_3 = \emptyset$). Actor $a_1$ communicates $C_{12} = \{s_1\}$ to $a_2$, who invalidates $s_1$ by presenting $I_{21}(s_1, s_2)$. Actor $a_3$, observing this interaction, updates their belief set based on the persuasion functions and update rules. This framework models the propagation of invalidation within a discursive network, capturing the dynamics of belief, communication, and influence. By formalizing these interactions, we can analyze and predict how invalidation affects the acceptance and rejection of statements among actors in the network.





\subsection{Modeling Discursive Networks}\label{sec:modeling}

\subsubsection{Single-Network Two-State Model}
\label{sec:single-network-two-state}

Let $n=|A|$ be the actor count in a collapsed single network with two mutually
exclusive statement states
\[
    S=\{r,f\},
\]
where $r$ denotes a valid or true state and $f$ denotes an invalid or false
state.  We work in proportions rather than raw counts.  The population state at
time $t$ is
\[
\boldsymbol{\pi}(t)=\bigl(\pi_T(t),\pi_F(t)\bigr)^{\mathsf T},\qquad
\pi_T(t)=\frac{T(t)}{n},
\qquad
\pi_F(t)=\frac{F(t)}{n}=1-\pi_T(t).
\]

The canonical model summarizes state change by two effective pressures:
\[
    a=\hbox{invalidation pressure from }r\hbox{ to }f,
    \qquad
    b=\hbox{correction pressure from }f\hbox{ to }r.
\]
There are two mathematically distinct interpretations.  In a one-round
Bernoulli model, $a$ and $b$ are transition probabilities and must lie in
$(0,1)$.  In a continuous-time hazard model, $a$ and $b$ are nonnegative rates
and may exceed one when measured per unit time.  Both interpretations have the
same stationary false-state share, but they have different time evolutions.
This distinction matters later because additive hazards such as
$q+(n-1)d$ can exceed one even though no one-round probability can.

In the discrete one-round interpretation, the population-level transition
matrix is
\begin{equation}\label{eq:single-network-system1}
    T_{a,b}=\begin{pmatrix}
    1-a & b\\
    a   & 1-b
    \end{pmatrix},
    \qquad
    \boldsymbol{\pi}(t+1)=T_{a,b}\boldsymbol{\pi}(t),
\end{equation}
and the false-state fraction obeys
\begin{equation}\label{eq:false-recursion-ab}
    \pi_F(t+1)=a+(1-a-b)\pi_F(t).
\end{equation}
In the continuous-time hazard interpretation, the corresponding scalar equation
is
\begin{equation}\label{eq:false-ode-ab}
    \frac{d}{dt}\pi_F(t)=a\bigl(1-\pi_F(t)\bigr)-b\pi_F(t)
    =a-(a+b)\pi_F(t).
\end{equation}

The mapping of this model to Definition~\ref{def:discursive_network} is provided
in Table~\ref{tab:Single-Network-Dual-State}.  The table states the base case
$a=p$, $b=q$; the next subsections retain the same structure while changing the
effective values of $a$ and $b$.

\begin{table}[htbp]
    \centering
    \setlength{\tabcolsep}{6pt}
    \renewcommand{\arraystretch}{1.15}
    \caption{Mapping of discursive network elements to the single-network two-state model.}
    \label{tab:Single-Network-Dual-State}
    \begin{tabularx}{6.5in}{@{}lX@{}}
    \toprule
    \textbf{Element in $N$} & \textbf{Instantiation in the single-network model} \\
    \midrule
    $S=\{r,f\}$ & Binary, mutually exclusive statement states: $r$ is valid/true and $f$ is invalid/false.\\
    $P_{ij}$ & Base persuasion or drift: $p$ contributes to $r\to f$ and $q$ contributes to $f\to r$.  In the generic form these are summarized by $a=p$ and $b=q$.\\
    $I_{ij}$ & In the single-network base case, invalidation is represented by the transition $r\to f$; correction is represented by $f\to r$.\\
    $C_{ij}$ & Message containing the current belief state $B_i\in\{\{r\},\{f\}\}$.\\
    $A$ & Fixed actor set; proportions refer to $n=|A|$.\\
    $B_i$ & Current belief of actor $a_i$: either $\{r\}$ or $\{f\}$.\\
    $U_j$ & Update rule applying either the one-round transition matrix $T_{a,b}$ or the continuous-time hazard equation with the same $a,b$ equilibrium.\\
    $G_i$ & Goal: persuade others to adopt actor $a_i$'s current belief.\\
    \bottomrule
    \end{tabularx}
\end{table}

\begin{lemma}[Single-network two-state equilibrium]
\label{lem:basic_invalidation}
Let $a,b>0$.  In the continuous-time hazard model
\eqref{eq:false-ode-ab}, the unique stationary false-state share is
\[
    \pi_F^*=\frac{a}{a+b},
    \qquad
    \pi_T^*=\frac{b}{a+b},
\]
and every solution satisfies
\[
    \pi_F(t)-\pi_F^*=e^{-(a+b)t}\bigl(\pi_F(0)-\pi_F^*\bigr).
\]
In the discrete one-round Bernoulli model \eqref{eq:single-network-system1}, if
$0<a,b<1$, then the same fixed point is
\[
\boldsymbol{\pi}^{*}
=
\begin{pmatrix}
\dfrac{b}{a+b}\\[6pt]
\dfrac{a}{a+b}
\end{pmatrix},
\]
and the chain converges geometrically from every initial distribution; the
second eigenvalue is $1-a-b$.  In the base case $a=p$, $b=q$, this gives
\[
\pi_T^*=\frac{q}{p+q},\qquad
\pi_F^*=\frac{p}{p+q}.
\]
\end{lemma}

\begin{proof}
For the continuous-time branch, setting the right-hand side of
\eqref{eq:false-ode-ab} to zero gives
$a-(a+b)\pi_F^*=0$, hence $\pi_F^*=a/(a+b)$ and
$\pi_T^*=1-\pi_F^*=b/(a+b)$.  Solving the linear differential equation gives
\[
    \pi_F(t)=\pi_F^*+e^{-(a+b)t}\bigl(\pi_F(0)-\pi_F^*\bigr).
\]
For the discrete branch, a fixed point satisfies
$\boldsymbol{\pi}=T_{a,b}\boldsymbol{\pi}$.  Writing
$\boldsymbol{\pi}^{\mathsf T}=(\pi_T,\pi_F)$ and using
$\pi_T+\pi_F=1$, the fixed-point equation reduces to
\[
    a\pi_T=b\pi_F,
\]
so $\pi_T=b/(a+b)$ and $\pi_F=a/(a+b)$.  The eigenvalues of $T_{a,b}$ are
$\mu_1=1$ and $\mu_2=1-a-b$.  Since $0<a,b<1$, we have
$0<a+b<2$ and therefore $|\mu_2|<1$, so deviations from equilibrium decay
geometrically:
\[
    \pi_F(t)-\pi_F^*=(1-a-b)^t\bigl(\pi_F(0)-\pi_F^*\bigr).
\]
\end{proof}

\paragraph{Interpretation of Lemma~\ref{lem:basic_invalidation}.}
The two-state model is governed by the ratio of inflow into the false state and
outflow from it.  The stationary false-state share is $a/(a+b)$; increasing the
invalidation pressure $a$ raises the long-run prevalence of $f$, while
increasing the correction pressure $b$ lowers it.  The base model is recovered
by setting $a=p$ and $b=q$.  The special case $p+q=1$ gives
$\boldsymbol{\pi}^{*}=(q,p)^{\mathsf T}$, but the model does not require
$p+q=1$.  When $a$ and $b$ are interpreted as probabilities, the transition
matrix entries must be nonnegative.  When they are interpreted as continuous-time
rates, the ODE branch applies and additive correction rates may exceed one
without creating negative transition probabilities.

\subsubsection{Single-Network Emergent Invalidation Model}
\label{sec:single-network-emergent}

The base two-state model separates false-state inflow from false-state outflow.
Emergent invalidation adds a fabrication or hallucination channel to the
ordinary truth-to-false drift.  We encode the combined false-production pressure
by
\begin{equation}\label{eq:a-single}
    a_{\mathrm{single}}=p+\lambda,
\end{equation}
where $p$ is intrinsic truth-to-false drift and $\lambda$ is fresh fabrication.
Equation~\eqref{eq:a-single} is the additive-rate convention and is also the
first-order approximation when $p$ and $\lambda$ are small Bernoulli
probabilities.  If they are modeled as independent one-round Bernoulli channels,
then replace Eq.~\eqref{eq:a-single} by the exact expression
\[
    a_{\mathrm{single}}=1-(1-p)(1-\lambda).
\]
The isolated false-to-true correction pressure is
\begin{equation}\label{eq:b-single}
    b_{\mathrm{single}}=q.
\end{equation}
The normalized state remains
\[
\boldsymbol{\pi}(t)=\bigl(\pi_T(t),\pi_F(t)\bigr)^{\mathsf T},
\qquad \pi_T(t)+\pi_F(t)=1,
\]
and the dynamics are
\begin{equation}\label{eq:single-lambda-matrix}
\boldsymbol{\pi}(t+1)=T_{\lambda}\boldsymbol{\pi}(t),\qquad
T_{\lambda}=\begin{pmatrix}
1-a_{\mathrm{single}} & q\\
a_{\mathrm{single}} & 1-q
\end{pmatrix}.
\end{equation}

The mapping of this specialization to Definition~\ref{def:discursive_network}
is provided in Table~\ref{tab:Single-Network-Invalidation}.

\begin{table}[htbp]
    \centering
    \setlength{\tabcolsep}{6pt}
    \renewcommand{\arraystretch}{1.15}
    \caption{Mapping of discursive-network elements to the single-network emergent-invalidation model.}
    \label{tab:Single-Network-Invalidation}
    \begin{tabularx}{6.5in}{@{}lX@{}}
        \toprule
        \textbf{Element in $N$} & \textbf{Instantiation in the emergent-invalidation model} \\
        \midrule
        $S=\{r,f\}$ & Binary statement states: $r$ is valid/true and $f$ is invalid/false.\\
        $P_{ij}$ & Intrinsic drift $p$ contributes to $r\to f$; correction $q$ contributes to $f\to r$.\\
        $I_{ij}$ & Internal invalidation and self-correction are summarized by the effective transition pressures $a_{\mathrm{single}}=p+\lambda$ and $b_{\mathrm{single}}=q$.\\
        $C_{ij}$ & Message from $a_i$ to $a_j$ carrying $B_i$.\\
        $A$ & Fixed actor set; proportions refer to $n=|A|$.\\
        $B_i$ & Current belief of actor $a_i$: $\{r\}$ or $\{f\}$.\\
        $U_j$ & Update rule applying $T_{\lambda}$ in Eq.~\eqref{eq:single-lambda-matrix}.\\
        $G_i$ & Goal: persuade other actors to adopt the current belief state.\\
        \bottomrule
    \end{tabularx}
\end{table}

\begin{lemma}[Single-network invalidation with fabrication]
\label{lem:single_lambda}
Let $a_{\mathrm{single}}=p+\lambda$ and $b_{\mathrm{single}}=q$.  Under
either the discrete one-round interpretation with
$0<a_{\mathrm{single}},b_{\mathrm{single}}<1$ or the continuous-time hazard
interpretation with $a_{\mathrm{single}},b_{\mathrm{single}}>0$, the
single-network emergent-invalidation model has stationary state
\[
\boldsymbol{\pi}^{*}
=
\begin{pmatrix}
\dfrac{q}{p+\lambda+q}\\[6pt]
\dfrac{p+\lambda}{p+\lambda+q}
\end{pmatrix}.
\]
In the discrete one-round branch, the second eigenvalue is
$1-(p+\lambda+q)$; in the continuous-time branch, the exponential convergence
rate is $p+\lambda+q$.
\end{lemma}

\begin{proof}
Apply Lemma~\ref{lem:basic_invalidation} with
$a=a_{\mathrm{single}}=p+\lambda$ and $b=b_{\mathrm{single}}=q$, using the
discrete or continuous branch according to the interpretation of the parameters.
\end{proof}

\paragraph{Interpretation.}
The fabrication term $\lambda$ increases the effective inflow into the false
state.  The isolated network is falsehood-majority exactly when
$p+\lambda>q$.  In the discrete branch, $p+\lambda+q<1$ gives monotone
geometric convergence and $p+\lambda+q\in(1,2)$ gives oscillatory geometric
convergence.  In the continuous-time branch, convergence is monotone and
exponential with rate $p+\lambda+q$.  Increasing $\lambda$ raises the false
equilibrium; it does not by itself imply slower convergence.

\subsubsection{Cross-Network Invalidation-Detection Model}
\label{sec:cross-network-detection}

Cross-network detection increases the effective false-to-true correction
pressure.  Consider a focal network $N_k$ with intrinsic false-production
pressure
\[
    a_k=p_k+\lambda_k
\]
under the additive-rate convention.  Let $q_k$ be its internal correction
pressure.  Let a partner network $N_j$ detect and correct false statements in
$N_k$ with external parameter $d_{jk}>0$.

There are two useful aggregation regimes.
\begin{enumerate}
\item \emph{Continuous-time hazard model.}  Independent correction clocks add,
      so the effective correction rate is
      \begin{equation}\label{eq:b-cross-rate}
      b_k^{\mathrm{rate}}=q_k+d_{jk}.
      \end{equation}
\item \emph{Exact one-round Bernoulli model.}  Internal and external correction
      channels both miss a false statement with probability
      $(1-q_k)(1-d_{jk})$, so the effective one-round correction probability is
      \begin{equation}\label{eq:b-cross-disc}
      b_k^{\mathrm{disc}}=1-(1-q_k)(1-d_{jk}).
      \end{equation}
\end{enumerate}
The proportion state is
\[
\boldsymbol{\pi}_k(t)=\bigl(\pi_{T,k}(t),\pi_{F,k}(t)\bigr)^{\mathsf T},
\qquad \pi_{T,k}(t)+\pi_{F,k}(t)=1,
\]
and the same two-state equilibrium applies with $a=a_k$: the continuous-time
ODE branch uses $b=b_k^{\mathrm{rate}}$, while the exact one-round Bernoulli
branch uses the transition matrix with $b=b_k^{\mathrm{disc}}$.

The mapping of this specialization to Definition~\ref{def:discursive_network}
is provided in Table~\ref{tab:Dual-Network}.

\begin{table}[htbp]
    \centering
    \setlength{\tabcolsep}{6pt}
    \renewcommand{\arraystretch}{1.15}
    \caption{Mapping of discursive-network elements to the cross-network invalidation-detection model.}
    \label{tab:Dual-Network}
    \begin{tabularx}{6.5in}{@{}lX@{}}
        \toprule
        \textbf{Element in $N_{1}\cup N_{2}$} & \textbf{Instantiation in the cross-network model} \\
        \midrule
        $S=\{r,f\}$ & Binary statement states shared by both networks: $r$ is valid/true and $f$ is invalid/false.\\
        $P_{ij}$ & Within $N_k$, $p_k$ contributes to $r\to f$ and $q_k$ contributes to $f\to r$. Across networks, detection by $N_j$ contributes $d_{jk}$ to false-to-true correction.\\
        $I_{ij}$ & Cross-network invalidation: a receiver in $N_j$ can identify a false state in $N_k$ and trigger correction with parameter $d_{jk}$.\\
        $C_{ij}$ & Message carrying the statement or belief state being evaluated.\\
        $A_k$ & Disjoint actor sets $A_1,A_2$; proportions refer to $n_k=|A_k|$.\\
        $B_{ij}$ & Belief of actor $a_{ij}\in A_k$: $\{r\}$ or $\{f\}$.\\
        $U_j$ & Update rule applying the two-state equilibrium with $a_k=p_k+\lambda_k$: the continuous-time hazard branch uses $b_k=q_k+d_{jk}$, and the exact Bernoulli branch uses the transition matrix with $b_k=1-(1-q_k)(1-d_{jk})$.\\
        $G_{ij}$ & Goal: persuade or correct other actors through communication and critique.\\
        \bottomrule
    \end{tabularx}
\end{table}

\begin{lemma}[Dual-network invalidation propagation]
\label{lem:expected_equilibrium}
Let a focal network have effective false-production parameter $a_k=p_k+\lambda_k$.
With one external detector $N_j$, the cross-network equilibrium is
\[
\pi_{F,k}^{\mathrm{cross,rate}}
=
\frac{p_k+\lambda_k}{p_k+\lambda_k+q_k+d_{jk}}
\]
in the continuous-time hazard model, and
\[
\pi_{F,k}^{\mathrm{cross,disc}}
=
\frac{p_k+\lambda_k}{p_k+\lambda_k+1-(1-q_k)(1-d_{jk})}
\]
in the exact one-round Bernoulli model.  The corresponding true-state shares are
$1-\pi_{F,k}^{\mathrm{cross,rate}}$ and
$1-\pi_{F,k}^{\mathrm{cross,disc}}$.
\end{lemma}

\begin{proof}
In the continuous-time hazard model, apply the ODE branch of
Lemma~\ref{lem:basic_invalidation} with $a=a_k$ and
$b=b_k^{\mathrm{rate}}=q_k+d_{jk}$.  This branch allows arbitrary positive
rates, so $q_k+d_{jk}$ need not be less than one.  In the exact one-round
Bernoulli model, apply the discrete branch with $a=a_k$ and
$b=b_k^{\mathrm{disc}}=1-(1-q_k)(1-d_{jk})$, assuming the resulting one-round
probabilities are admissible.  Both substitutions give the displayed fixed
points.
\end{proof}

\paragraph{Interpretation of Lemma~\ref{lem:expected_equilibrium}.}
The cross-network equilibrium depends on the ratio of false-production pressure
to total correction pressure.  Increasing $\lambda_k$ or $p_k$ raises the
false-state share.  Increasing $q_k$ or $d_{jk}$ lowers it.  In the rate model,
internal and external correction add linearly.  In the exact one-round Bernoulli
model, their combined effect saturates below one, as probabilities must.

\subsubsection{Single- vs.\ Cross-Network Models with Invalidation}
\label{sec:single-cross-comparison}

For an isolated single network under the additive-rate convention,
Lemma~\ref{lem:single_lambda} gives
\[
\pi_F^{\mathrm{single}}
=
\frac{p+\lambda}{p+\lambda+q},
\qquad
\pi_T^{\mathrm{single}}
=
\frac{q}{p+\lambda+q}.
\]
With one external detector, Lemma~\ref{lem:expected_equilibrium} gives
\[
\pi_F^{\mathrm{cross,rate}}
=
\frac{p+\lambda}{p+\lambda+q+d}
\]
in the continuous-time hazard model, and
\[
\pi_F^{\mathrm{cross,disc}}
=
\frac{p+\lambda}{p+\lambda+1-(1-q)(1-d)}
\]
in the exact one-round Bernoulli model.

\begin{lemma}[Cross-network detection lowers falsehood prevalence]
\label{lem:cross_vs_single}
If $d>0$, then
\[
\pi_F^{\mathrm{cross,rate}}<\pi_F^{\mathrm{single}}
\]
in the continuous-time hazard model, and
\[
\pi_F^{\mathrm{cross,disc}}<\pi_F^{\mathrm{single}}
\]
in the exact one-round Bernoulli model.
\end{lemma}

\begin{proof}
In the isolated model the correction parameter is $b=q$.  With one external
detector, the correction parameter is $q+d>q$ in the rate model and
$1-(1-q)(1-d)=q+d-qd>q$ in the exact Bernoulli model.  Since $a/(a+b)$ is
strictly decreasing in $b$ for $a>0$, the result follows.
\end{proof}

\paragraph{Interpretation.}
External scrutiny lowers the steady-state falsehood share by increasing the
false-to-true correction pressure.  The improvement is monotone in $d$.  The
continuous-time hazard model permits additive accumulation of rates, whereas the
one-round Bernoulli model saturates because total correction probability cannot
exceed one.

\subsubsection{How Many Agents Guarantee a Target Falsehood Level?}
\label{sec:agent-requirements}

Let $a=p+\lambda$ denote the effective false-production parameter of a focal
network.  Suppose the focal network has internal correction parameter $q$ and is
coupled to $n-1$ partner networks, each supplying external correction parameter
$d$.

\begin{lemma}[Effective correction: discrete probability versus continuous hazard]
\label{lem:effective_hazard}
The effective correction parameter is
\[
    b_n^{\mathrm{disc}}=1-(1-q)(1-d)^{n-1}
\]
in the exact one-round Bernoulli model, and
\[
    b_n^{\mathrm{rate}}=q+(n-1)d
\]
in the continuous-time hazard model.  The additive expression is exact for
continuous-time rates and is the first-order small-probability approximation to
the discrete expression.
\end{lemma}

\begin{proof}
In one discrete round, all external agents miss the false statement with
probability $(1-d)^{n-1}$, and the internal self-repair mechanism misses with
probability $(1-q)$.  Therefore all channels miss with probability
$(1-q)(1-d)^{n-1}$, giving
$b_n^{\mathrm{disc}}=1-(1-q)(1-d)^{n-1}$.  In continuous time, the minimum of
independent exponential clocks is exponential with rate equal to the sum of the
component rates, so $b_n^{\mathrm{rate}}=q+(n-1)d$.
\end{proof}

\begin{proposition}[Agent requirement for a tolerance $\varepsilon$]
\label{prop:n_agents}
Let $\varepsilon\in(0,1)$ and define
\[
    B_\varepsilon=a\left(\frac1\varepsilon-1\right).
\]
In the exact one-round Bernoulli model,
\[
    \pi_f^{(n),\mathrm{disc}}
    =
    \frac{a}{a+1-(1-q)(1-d)^{n-1}}.
\]
The minimum number of coupled networks required to guarantee
$\pi_f^{(n),\mathrm{disc}}\le\varepsilon$ is
\[
 n_{\min}^{\mathrm{disc}}
 =
 \begin{cases}
 1, & B_\varepsilon\le q,\\[4pt]
 1+\left\lceil
 \dfrac{\log\left((1-B_\varepsilon)/(1-q)\right)}{\log(1-d)}
 \right\rceil,
 & q<B_\varepsilon<1,\\[12pt]
 \text{unattainable by finite one-round Bernoulli detection}, & B_\varepsilon\ge1.
 \end{cases}
\]
In the continuous-time hazard model,
\[
    \pi_f^{(n),\mathrm{rate}}
    =
    \frac{a}{a+q+(n-1)d},
\]
and
\[
    n_{\min}^{\mathrm{rate}}
    =
    \max\left\{1,
    \left\lceil
    1+\frac{a(1/\varepsilon-1)-q}{d}
    \right\rceil
    \right\}.
\]
\end{proposition}

\begin{proof}
For the exact Bernoulli model, the discrete branch of Lemma~\ref{lem:basic_invalidation} gives the false-state fixed point $a/(a+b_n^{\mathrm{disc}})$.  For the continuous-time hazard model, the ODE branch gives $a/(a+b_n^{\mathrm{rate}})$ for any positive rates.  In the discrete
case, $a/(a+b_n^{\mathrm{disc}})\le\varepsilon$ is equivalent to
$b_n^{\mathrm{disc}}\ge B_\varepsilon$.  If $B_\varepsilon\le q$, the isolated
network already meets the target.  If $q<B_\varepsilon<1$, solve
\[
1-(1-q)(1-d)^{n-1}\ge B_\varepsilon,
\]
which gives
\[
(1-d)^{n-1}\le\frac{1-B_\varepsilon}{1-q}.
\]
Since $\log(1-d)<0$, this yields the stated ceiling formula.  If
$B_\varepsilon\ge1$, no finite number of one-round Bernoulli detectors can
achieve the required correction probability.  The continuous-time formula is
obtained by substituting $b_n^{\mathrm{rate}}=q+(n-1)d$ and solving the same
inequality for $n$.
\end{proof}

\paragraph{Interpretation.}
The discrete and continuous formulas answer different engineering questions.
The exact Bernoulli model describes one round in which each partner has bounded
detection probability.  The continuous-time model describes independent
detection clocks acting over time; rates can add without violating a one-round
probability cap.  The notation in simulations and figures should state which
regime is being used.

\subsection{FOO Algorithm with Integrity Verification}\label{sec:foo-algorithm}
The asymmetry exploited by the FOO algorithm is the conditional verification
channel of Theorem~\ref{thm:verification-channel}.  FOO couples an arbitrary
ensemble of LLM agents, tool-using processes, human reviewers, or formal
checkers to a critique-and-harmonization loop.  The algorithm reduces delivered
invalidity when critics have positive sensitivity on non-common-mode errors,
false-positive burden is acceptable, and flagged invalid outputs are repaired or
rejected.  This framing connects FOO to debate, process supervision, verifier
training, tool use, and mixture-of-agents architectures
\parencite{irving2018debate,du2023multiagentdebate,cobbe2021trainingverifiers,lightman2023verify,schick2023toolformer,wang2024moa}.

The FOO protocol also requires trust in the integrity of the interaction log. In
collaborative scientific work, the provenance of each response, critique,
harmonization decision, and revision becomes essential for reproducibility and
accountability.  We therefore extend the basic protocol with tamper-evident
logging based on signed hash links.

\subsubsection{Core FOO Protocol}
\begin{figure}[htbp]
    \centering
    \begin{tikzpicture}[node distance=1.5cm and 2cm, >=Stealth, thick, scale=0.85] 
        \tikzstyle{block} = [rectangle, draw, text centered, rounded corners, minimum height=1.2cm]
        \tikzstyle{arrow} = [->, thick]
        \tikzstyle{circle} = [circle, fill, inner sep=1.5pt]
        \tikzstyle{decision} = [diamond, draw, text centered, minimum width=2cm, minimum height=1.5cm, inner sep=0pt]
        \tikzstyle{stop} = [rectangle, draw, text centered, rounded corners, minimum height=1.2cm, fill=red!30]

        \node (request) [block] {Request};
        \node (cir) [block, below of=request, yshift=-0.5cm] {Distribute};

        \node (llm1) [block, below of=cir, xshift=-3cm, yshift=-1cm] {LLM$_1$};
        \node (dots1) [block, right of=llm1] {...};
        \node (llmj) [block, right of=dots1] {LLM$_j$};
        \node (dots2) [block, right of=llmj] {...};
        \node (llmn) [block, right of=dots2] {LLM$_n$};
        
        \node (o1) [block, below of=llm1, yshift=-1cm] {$O_1$};
        \node (dots3) [block, right of=o1] {...};
        \node (oj) [block, below of=llmj, yshift=-1cm] {$O_j$};
        \node (dots4) [block, right of=oj] {...};
        \node (on) [block, below of=llmn, yshift=-1cm] {$O_n$};
        
        \node (concat) [block, below of=dots3, xshift=1.5cm, yshift=-1cm] {Assemble};
        \node (decision) [decision, below of=concat, yshift=-0.5cm] {Differ?};
        \node (stop) [stop, right of=decision, xshift=2cm] {Respond};
        \node (consensus) [block, below of=decision, yshift=-1cm] {Add Consensus Instructions};
        
        \draw [arrow] (request) -- (cir);
        \draw [arrow] (cir) -- (llm1);
        \draw [arrow] (cir) -- (dots1);
        \draw [arrow] (cir) -- (llmj);
        \draw [arrow] (cir) -- (dots2);
        \draw [arrow] (cir) -- (llmn);
        
        \draw [arrow] (llm1) -- (o1);
        \draw [arrow] (dots1) -- (dots3);
        \draw [arrow] (llmj) -- (oj);
        \draw [arrow] (dots2) -- (dots4);
        \draw [arrow] (llmn) -- (on);
        
        \draw [arrow] (o1) -- (concat);
        \draw [arrow] (oj) -- (concat);
        \draw [arrow] (on) -- (concat);
        \draw [arrow] (dots3) -- (concat);
        \draw [arrow] (dots4) -- (concat);
        
        \draw [arrow] (concat) -- (decision);
        \draw [arrow] (decision) -- node[anchor=west] {Yes} (consensus);
        
        \draw [arrow] (consensus.west) -- ++(-2,0) -- ++(0,12) -- (cir.west);
        \draw [arrow] (decision.east) -- node[anchor=south] {No} (stop.west);
        
    \end{tikzpicture}

    \caption{FOO consensus loop.  
             An arbitrary set of agents \(a_{1},\dots,a_{m}\) receives
             the user task, produces candidate answers, cross-critiques
             peers, and feeds all critiques to one or more harmonizers
             \(h\).  Harmonizers synthesise the feedback; agents revise
             and the loop continues until a convergence criterion is
             met.  The design allows any number or type of agents and
             any custom instruction set, making the architecture task-
             and model-agnostic.}
    \label{fig:consensus-flow}
\end{figure}

The protocol has four phases:

1. \emph{Broadcast}: an initial user task is broadcast to every
   active agent.  Each agent returns a first-pass answer.

2. \emph{Cross-examination (FOO step)}: every agent receives the instruction
   \emph{``find the flaws in …"} followed by \emph{all} peer answers
   except its own, and produces a critique.  
   This implements the cross-detection hazard: an error overlooked by
   one model is likely to be flagged by at least one other.

3. \emph{Harmonization}: one or more agents are flagged as
   \emph{harmonizers}.  They aggregate the entire set of critiques,
   separate agreements from contradictions, and emit a structured
   ``judgement."  Harmonizers can use any rubric-majority vote, weighted
   confidence, specialist veto, to convert divergent feedback into a
   common set of observations.

4. \emph{Revision and loop}: every non-harmonizer ingests the judgement and
   regenerates its answer, optionally rebutting points it believes to be
   wrong.  The cycle repeats until a termination condition is met
   (identical outputs, bounded edit distance, or a maximum number of
   rounds).  The final harmonizer synthesis is returned to the user.

Because the agents; instructions; stopping rule; and comparison metric
are all configurable, the same code base supports tasks as different as
mathematical proof sketching, literature surveying, or code review.  The
FOO loop thus acts as a \emph{versatile wrapper} that upgrades solitary
generation into a networked, self-auditing process, realising in
practice the external detection hazard that pushes the system into the
truth-dominant regime predicted by the theory.

\begin{algorithm}
\caption{FOO with integrity logging}
\label{algo:foo-blockchain}
\begin{algorithmic}[1]
\Require user task $T$; agent set $\mathcal{A}$; convergence test
\Ensure final harmonized answer $R$ with verified interaction log
\State $\mathcal{B} \gets$ initialize blockchain with genesis block
\State broadcast $T$ to every $a\in\mathcal{A}$ and collect initial answers
\State \textbf{log} initial responses to blockchain $\mathcal{B}$
\Repeat
    \ForAll{agent $a\in\mathcal{A}$}
        \State supply $a$ with "find flaws in" $\{\text{answers of } \mathcal{A}\setminus\{a\}\}$
        \State receive critique $C_{a}$
        \State \textbf{log} critique $C_a$ to blockchain $\mathcal{B}$
    \EndFor
    \State harmonizer(s) $h$ aggregate $\{C_{a}\}$ into judgement $J$
    \State \textbf{log} harmonization decision to blockchain $\mathcal{B}$
    \ForAll{non-harmonizer $a$}
        \State regenerate answer $A_{a}$ conditioned on $J$
        \State \textbf{log} revision to blockchain $\mathcal{B}$
    \EndFor
\Until{convergence test satisfied}
\State \Return final $J$ and verified blockchain $\mathcal{B}$
\end{algorithmic}
\end{algorithm}


\subsubsection{Tamper-evident logging}
\label{sec:blockchain}

Each FOO interaction can be recorded in a signed hash-linked log in the sense of
Haber and Stornetta's time-stamping construction \parencite{haber1991timestamp}.
This construction should be called a \emph{tamper-evident log}; it becomes a
blockchain only if the implementation also specifies a distributed-consensus or
external-anchoring mechanism.  Each entry has the form
\[
E_i=(m_i,t_i,\mathrm{id}_i,\sigma_i,h_i),
\qquad
h_i=H(h_{i-1}|m_i|t_i|\mathrm{id}_i|\sigma_i),
\]
where $H$ is a collision-resistant hash function such as SHA-256, and
$\sigma_i$ is a digital signature, for example an EdDSA signature, over
$(m_i,t_i,h_{i-1})$ under agent $\mathrm{id}_i$'s long-term signing key.

The log gives tamper evidence: changing an old entry changes every subsequent
hash.  It also gives per-entry non-repudiation when signatures and key
management are implemented correctly.  It does not by itself prevent an
adversary who controls the storage substrate from rewriting the log forward from
a modified entry and recomputing every later hash.  Stronger tamper resistance
requires anchoring the current head $h_N$ to an external append-only medium,
such as an RFC~3161 trusted time-stamping authority, a public-chain commitment,
or a Byzantine-fault-tolerant consensus service \parencite{rfc3161,castro1999pbft}.

The integrity layer adds four checkpoint types to Algorithm~\ref{algo:foo-blockchain}:
\begin{enumerate}
\item initial response logging after broadcast,
\item critique logging during cross-examination,
\item harmonization-decision logging,
\item revision logging during iteration.
\end{enumerate}
A security claim should specify the threat model: who holds signing keys, what
happens if an agent key is compromised, whether timestamps are trusted, whether
log heads are externally anchored, and whether the system promises only
tamper-evidence or also tamper-resistance under a consensus protocol.

\section{Theoretical Validation and Parameter Analysis}
\label{sec:theoretical-validation}

This section illustrates the dynamics of the discursive-network
model.  The purpose is not to claim empirical calibration of the latent hazards
\(p,q,\lambda,d\).  Rather, the purpose is to verify internal consistency:
the model should preserve the population simplex, converge to the fixed points
proved in Section~\ref{sec:modeling}, and distinguish clearly between the exact
one-round Bernoulli regime and the continuous-time rate regime.

The literature cited below motivates the qualitative scale of the parameters,
but it does not directly estimate them.  Reported hallucination rates,
self-evaluation scores, process-supervision gains, and verifier-performance
numbers are task-level outcomes under specific benchmarks; they are not
plug-in estimates of the hazards in Lemmas~\ref{lem:basic_invalidation},
\ref{lem:single_lambda}, and~\ref{lem:expected_equilibrium}.  Thus the values in
Table~\ref{tab:param-calibration} should be read as an illustrative
parameterization for a reproducible regime check.  A future empirical study
should estimate these hazards directly from controlled experiments that
separately measure false production, intrinsic correction, external detection,
false-positive review burden, repair success, and common-mode blind spots.

\subsection{Illustrative parameter specification}
\label{sec:illustrative-parameters}

Published studies motivate an invalidation-dominant single-agent regime without
directly determining the hazards.  Medical-domain evaluations and hallucination
mitigation studies document that factual errors can remain substantial in
specialized settings \parencite{sallam2023chatgpt,thirunavukarasu2023large,Ji2023SelfReflection},
while \textcite{Zhang2024SelfAlignment} emphasize that intrinsic
self-evaluation requires calibration and remains weak in some factuality
settings.  Work on learned verifiers and process supervision shows that
externalized checking can improve mathematical reasoning, but the gains depend
on the verifier, the task, and the available evidence
\parencite{cobbe2021trainingverifiers,lightman2023verify}.  Conversely,
intrinsic self-correction without external feedback can fail or even degrade
reasoning performance \parencite{huang2023cannotselfcorrect}.  These findings
support exploring regimes in which single-agent error production exceeds
intrinsic repair while cross-agent or tool-supported verification provides an
additional correction channel.

Table~\ref{tab:param-calibration} fixes one such regime.  We write
\[
    a := p+\lambda
\]
for the effective true-to-false drift in the additive mean-field convention.
For the numerical values below, \(a=0.075\).  If \(p\) and \(\lambda\) are
instead interpreted as independent one-round Bernoulli channels, the exact
combined drift is \(1-(1-p)(1-\lambda)=0.0739\); the difference is
\(p\lambda=0.0011\).  The figures use the additive convention so that they
match the closed-form expressions in the preceding lemmas.

\begin{table}[htbp]
    \centering
    \setlength{\tabcolsep}{5pt}
    \renewcommand{\arraystretch}{1.15}
    \caption{Illustrative parameter values for the corrected dynamics.
    The values are chosen to display the qualitative regimes of the theory;
    they are not fitted estimates.  The same numerical value \(d=0.19\) is used
    as a one-round Bernoulli detection probability in the Monte-Carlo plots and
    as a continuous-time rate in the analytic rate curve; those are separate
    interpretations used for comparison.}
    \label{tab:param-calibration}
    \begin{tabularx}{\linewidth}{@{}lclX@{}}
        \toprule
        Symbol & Value & Model role & Interpretation in this section \\
        \midrule
        \(p\) & \(0.02\) & intrinsic true-to-false drift &
        Small background slippage from valid to invalid state. \\[2pt]
        \(q\) & \(0.05\) & intrinsic false-to-true repair &
        Modest self-repair, reflecting that intrinsic self-evaluation is
        useful but unreliable without external evidence. \\[2pt]
        \(\lambda\) & \(0.055\) & additional false-production drift &
        Chosen with \(p,q\) so that the isolated network has false equilibrium
        \(f_{\mathrm{single}}^*=0.60\), an invalidation-dominant baseline. \\[2pt]
        \(d\) & \(0.19\) & per-peer external detection/correction &
        Chosen to illustrate a positive verification channel: one external
        peer lowers the false equilibrium to approximately \(0.24\), and a
        nine-agent continuous-rate system falls below a \(5\%\) target. \\
        \bottomrule
    \end{tabularx}
\end{table}

\subsection{Analytic consistency checks}
\label{sec:analytic-consistency-checks}

The corrected two-state dynamics aggregate false production into \(a\) and
false correction into \(b\).  In the exact one-round Bernoulli model,
\[
    f_{t+1}=a+(1-a-b)f_t,
    \qquad
    f^*=\frac{a}{a+b}.
\]
In the continuous-time rate model,
\[
    \dot f=a(1-f)-bf,
    \qquad
    f^*=\frac{a}{a+b}.
\]
The same equilibrium formula appears in both branches, but the admissible
interpretation of \(b\) differs.  In the Bernoulli branch \(b\le 1\) is a
probability.  In the rate branch \(b\) can exceed \(1\), because it is a hazard
rate in the differential equation rather than an entry of a stochastic matrix.

For the isolated network, \(b=q=0.05\).  Hence
\[
    f_{\mathrm{single}}^*
    =
    \frac{p+\lambda}{p+\lambda+q}
    =
    \frac{0.075}{0.125}
    =
    0.60,
    \qquad
    r_{\mathrm{single}}^*=0.40.
\]
The isolated network is therefore invalidation-dominant in this illustrative
regime.

\begin{figure}[htbp]
    \centering
    \begin{subfigure}[b]{0.48\textwidth}
        \centering
        \includegraphics[width=\linewidth]{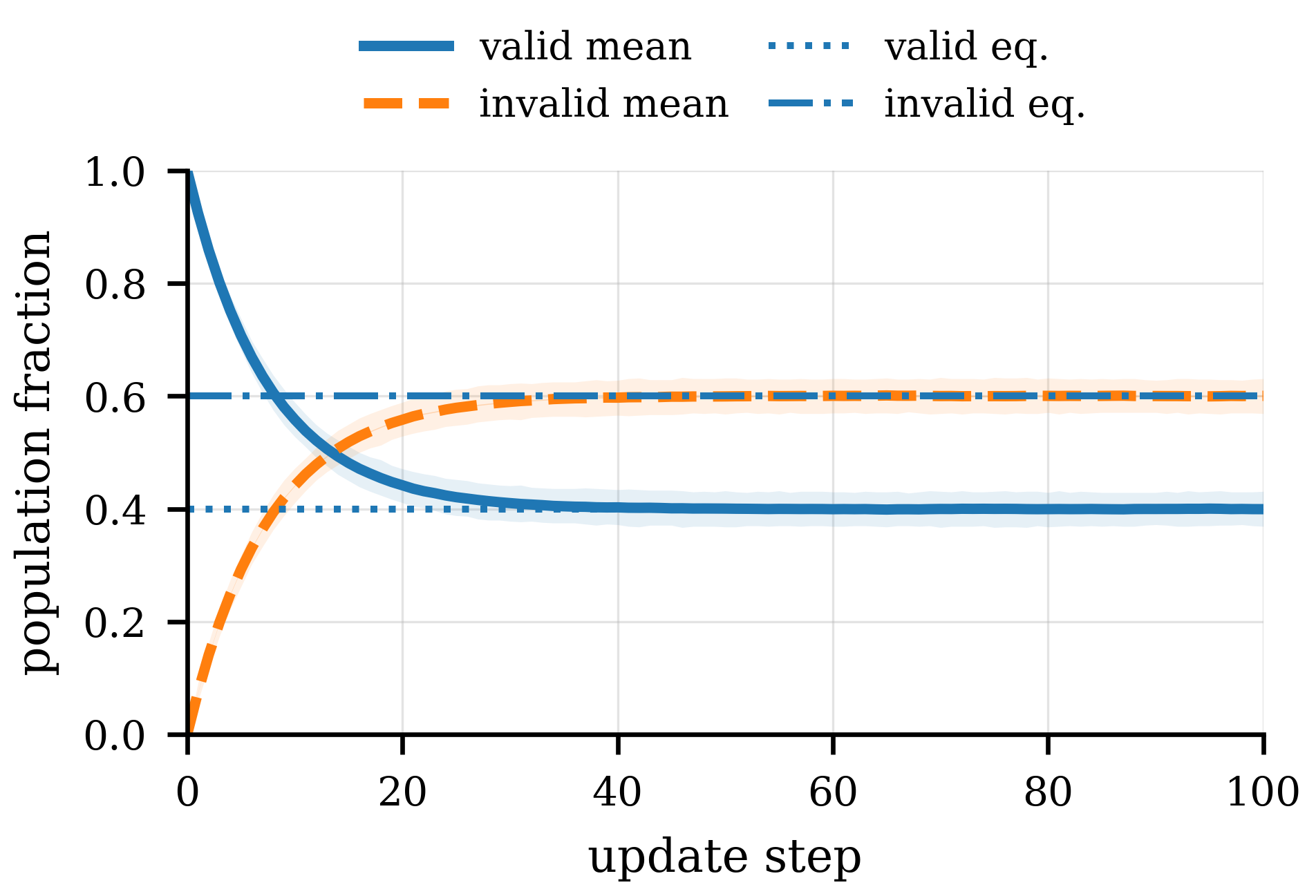}
        \caption{}
        \label{fig:single-net}
    \end{subfigure}
    \hfill
    \begin{subfigure}[b]{0.48\textwidth}
        \centering
        \includegraphics[width=\linewidth]{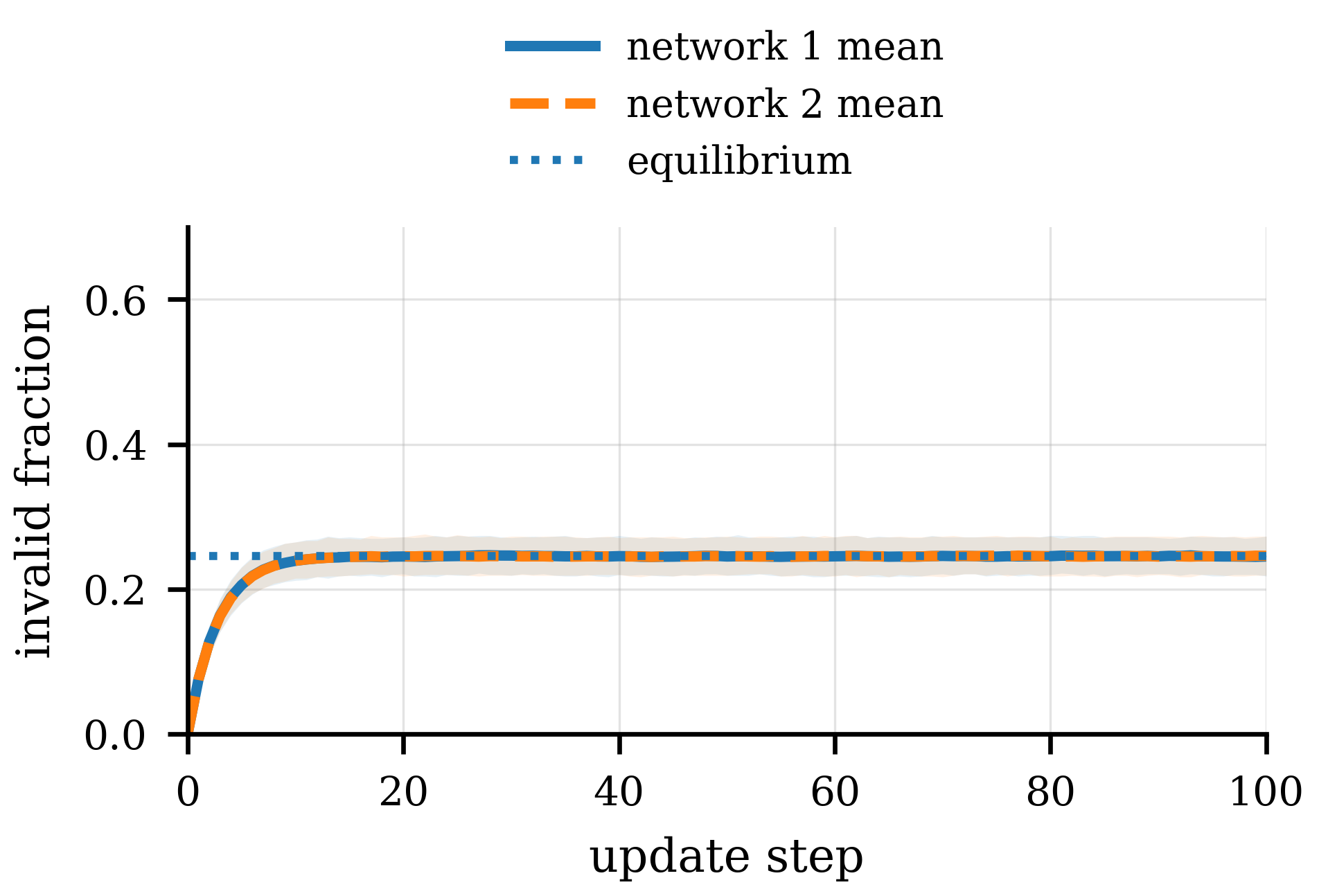}
        \caption{}
        \label{fig:dual-net}
    \end{subfigure}

    \caption{\textbf{(a)} Single-network two-state dynamics with
    \(a=p+\lambda=0.075\) and \(b=q=0.05\).  The trajectories converge to
    \(f_{\mathrm{single}}^*=0.60\) and \(r_{\mathrm{single}}^*=0.40\), as
    predicted by Lemma~\ref{lem:single_lambda}.
    \textbf{(b)} Two identical networks with one external detection channel.
    The simulation uses the exact Bernoulli correction probability
    \(b=1-(1-q)(1-d)=0.2305\), so both networks converge to the same false
    equilibrium \(f_{\mathrm{cross,disc}}^*\approx0.2455\).  Any persistent
    asymmetry between two identical networks would indicate non-equilibration,
    a code--equation mismatch, or a different implemented update rule.}
    \label{fig:single-vs-dual}
\end{figure}

For one external peer, the two interpretations are close but not identical.
The continuous-rate or small-probability approximation uses
\[
    b_{\mathrm{cross,rate}}=q+d=0.24,
    \qquad
    f_{\mathrm{cross,rate}}^*
    =
    \frac{0.075}{0.075+0.05+0.19}
    \approx 0.2381.
\]
The exact one-round Bernoulli model combines intrinsic and external correction
as independent chances to detect the same false statement:
\[
    b_{\mathrm{cross,disc}}
    =
    1-(1-q)(1-d)
    =
    1-(0.95)(0.81)
    =
    0.2305,
\]
so that
\[
    f_{\mathrm{cross,disc}}^*
    =
    \frac{0.075}{0.075+0.2305}
    \approx 0.2455.
\]
Both values are substantially below the isolated value \(0.60\).  The exact
Bernoulli Monte-Carlo simulation in Figure~\ref{fig:single-vs-dual} uses
\(b_{\mathrm{cross,disc}}\); the continuous-rate value is reported as the
corresponding ODE approximation.

\subsection{Simulation protocol}
\label{sec:simulation-protocol}

The finite-population simulations use the stochastic update that corresponds
directly to the corrected two-state recursion.  Let \(N\) be the number of
claims or actors being tracked in a network and let \(F_t\) denote the number
in the invalid state at step \(t\).  Given \(F_t\), define
\[
    X_t \sim \operatorname{Binomial}(N-F_t,a),
    \qquad
    C_t \sim \operatorname{Binomial}(F_t,b),
\]
where \(X_t\) is the number of valid items that become invalid and \(C_t\) is
the number of invalid items corrected.  The update is
\[
    F_{t+1}=F_t+X_t-C_t,
    \qquad
    T_{t+1}=N-F_{t+1}.
\]
This construction preserves \(T_t+F_t=N\) exactly for every sample path.
For the single-network figure, \(b=q\).  For the two-network figure, each false
item survives a round only if it escapes both intrinsic repair and the external
peer, so
\[
    b=1-(1-q)(1-d).
\]
The plotted confidence bands are pointwise \(95\%\) Monte-Carlo intervals over
independent runs.  The accompanying script
\texttt{FOO\_generate\_plots.py} uses \(N=1000\), \(1000\) runs, \(100\) update
steps, and a fixed random seed by default.  It writes vector PDF files for the
paper and high-resolution PNG files for quick inspection.

\subsection{Agent count under Bernoulli and rate regimes}
\label{sec:agent-count-parameters}

For \(n\) agents, the exact one-round Bernoulli correction probability is
\[
    b_n^{\mathrm{disc}}
    =
    1-(1-q)(1-d)^{n-1},
\]
whereas the continuous-time rate model uses
\[
    b_n^{\mathrm{rate}}
    =
    q+(n-1)d.
\]
The two formulas agree to first order when \(q\) and \(d\) are small, but they
differ qualitatively for large \(n\).  The Bernoulli correction probability is
capped by \(1\); the rate \(q+(n-1)d\) is unbounded because it is the total
hazard of independent exponential clocks.

For a target false share \(\varepsilon=0.05\), the equilibrium condition
\(f^*=a/(a+b)\le\varepsilon\) is equivalent to
\[
    b \ge a\left(\frac{1}{\varepsilon}-1\right).
\]
With \(a=0.075\),
\[
    a\left(\frac{1}{0.05}-1\right)=0.075(19)=1.425.
\]
No one-round Bernoulli system can achieve this target under the illustrative
parameters, because \(b_n^{\mathrm{disc}}\le 1\) for every \(n\).  Its limiting
false share is
\[
    \lim_{n\to\infty}
    \frac{a}{a+b_n^{\mathrm{disc}}}
    =
    \frac{0.075}{0.075+1}
    \approx0.0698
    >
    0.05.
\]
Thus the exact Bernoulli model has a nonzero floor above the chosen tolerance.

The continuous-rate model gives a different design calculation:
\[
    f_n^{\mathrm{rate}}
    =
    \frac{a}{a+q+(n-1)d}.
\]
Solving \(f_n^{\mathrm{rate}}\le0.05\) yields
\[
    n_{\min}^{\mathrm{rate}}
    =
    \left\lceil
    1+\frac{a(1/0.05-1)-q}{d}
    \right\rceil
    =
    \left\lceil
    1+\frac{1.425-0.05}{0.19}
    \right\rceil
    =
    9.
\]
Indeed,
\[
    f_8^{\mathrm{rate}}\approx0.0515>0.05,
    \qquad
    f_9^{\mathrm{rate}}\approx0.0456<0.05.
\]
The statement ``nine agents suffice'' is therefore a statement about the
continuous-time additive-hazard model, not about exact one-round Bernoulli
detection.

\begin{figure}[htbp]
    \centering
    \includegraphics[width=0.74\linewidth]{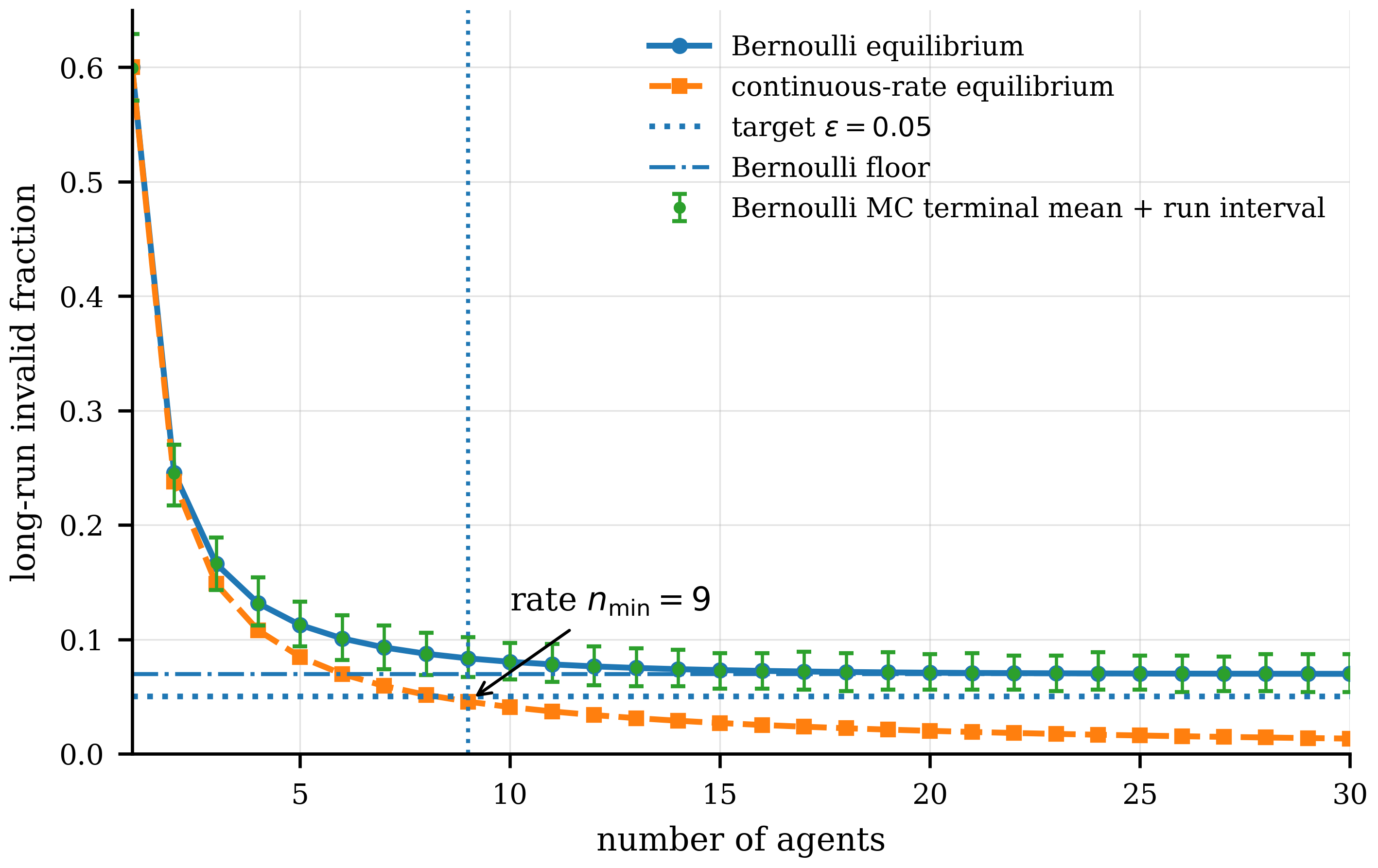}
    \caption{Long-run false share as a function of the number of agents.
    The exact Bernoulli curve uses
    \(b_n^{\mathrm{disc}}=1-(1-q)(1-d)^{n-1}\) and approaches the floor
    \(0.075/(0.075+1)\approx0.0698\), which remains above the \(5\%\) target.
    The continuous-rate curve uses
    \(b_n^{\mathrm{rate}}=q+(n-1)d\) and crosses the \(5\%\) target at
    \(n=9\).  The figure therefore displays two different regimes, not two
    simulations of the same stochastic matrix.}
    \label{fig:agent-count}
\end{figure}

\subsection{Interpretation}
\label{sec:parameter-interpretation}

The numerical lesson is that external verification can strongly reduce the
false equilibrium, but the strength of the conclusion depends on the correction
regime.  A single external peer lowers the illustrative false share from
\(0.60\) to about \(0.24\).  However, the exact one-round Bernoulli model still
has a probability cap; under the chosen values it cannot reach a \(5\%\) target
no matter how many same-strength one-round detectors are added.  The
continuous-time model reaches that target with nine agents because independent
exponential correction hazards add without a probability cap.

Accordingly, these plots should be described as consistency checks and regime
illustrations.  They do not by themselves validate the hazards empirically.  A
research-quality empirical validation would need to estimate \(p,q,\lambda,d\)
and, for the verification-channel theorem, \((\delta_m,\eta_m,c,\rho)\) from
task-specific experiments.  The figures generated by
\texttt{FOO\_generate\_plots.py} are reproducibility artifacts for the corrected
mathematics: equations, code, and captions all implement the same two-state
model.

\section{Discussion}\label{sec:discussion}
\subsection{Ethical Concerns}
\label{sec:ethics}
The first and most important observation regarding discursive networks with LLMs is ethical: there is a risk of transgression when the human(s) in a discursive network use it as a means to oursource reasoning instead of deploying it as augmented intelligence. 
In linguistics, \emph{epithesis} refers to the addition of a sound or
letter to the end of a word without changing its meaning.  By analogy we
identify an ethical concern in discursive networks: \textbf{epithesis}, wherein an individual seeks authorship on an artifact to
which they have contributed only superficial edits, or none at all.
Like the linguistic phenomenon, the intervention leaves the substantive
content untouched while appending an external element that alters
perception rather than substance.  Epithesis does not meet
the formal threshold of plagiarism, yet it belongs to the same family of
misappropriations because it places a symbolic layer of credit ``upon''
the discourse without engaging in its intellectual construction.  In the
context of discursive networks this behaviour distorts the link between
contribution and attribution, undermining the very mechanism of
cross-agent validation that the network is designed to support.

Authorship norms present another axis of concern.  As dozens of agents
contribute micro-edits, intellectual responsibility becomes
increasingly opaque, complicating both credit assignment and error
tracing.  Empirical work across disciplines shows that diffuse
contributions encourage honourary or ``gift" authorship, diluting
accountability and undermining public trust in published findings
\cite{maruvsic2011systematic}. What constitutes authorship in discursive network is an open question that will take time to settle.  

The integrity of discursive networks fundamentally depends on the ability to verify the authenticity and provenance of each contribution, whether from human or artificial agents. When scientific conclusions emerge from iterative exchanges among multiple participants, traditional notions of authorship become complicated by the distributed nature of intellectual labor and the possibility of post-hoc modification of interaction records. This challenge is particularly acute in combating epithesis, as the practice thrives in environments where genuine contributions cannot be distinguished from superficial additions or retroactive claims of involvement. A robust solution requires cryptographic mechanisms that create tamper-evident logs of all interactions, making it computationally infeasible to fabricate authorship claims after the fact. By implementing blockchain-based integrity verification for agent communications, discursive networks can establish a trustworthy foundation where each participant's actual contributions are permanently recorded and verifiable. This technical infrastructure does more than prevent fraud; it creates positive incentives for meaningful engagement by ensuring that substantial intellectual contributions receive proper attribution while making epithetic behavior both detectable and reputationally costly. The result is a research environment where collaborative human-AI knowledge production can proceed with confidence in the integrity of the underlying interaction records.

The energy footprint of large
discursive networks pose another ethical dilemma.  Each critical review involves at least one
forward-and-backward pass through a language model, so if a manuscript is
critiqued by a \emph{single} external agent the computational cost grows
\emph{linearly} with the number of agents, \(E(N)=\Theta(N)\).  When
every agent critiques every other agent (the fully connected case that is
ideal for robustness) the number of pairwise exchanges scales as
\(N(N-1)/2=\Theta(N^{2})\), and so does the energy consumption.
Moreover, once those quadratic interactions have occurred the network
typically runs a consensus or ``harmonization" phase to reconcile
conflicting edits; common distributed algorithms complete in
\(O(\log N)\) synchronous rounds.  The aggregate budget therefore climbs
to \(\Theta \bigl(N^{2}\log N\bigr)\) for end-to-end validation,
dwarfing the cost of the original single-agent composition.  In an era
when each large-model inference already carries a measurable carbon
footprint, the quadratic-plus overhead raises difficult questions about
the sustainability of scaling discursive networks without parallel
investment in greener compute or more frugal validation protocols.

Because those costs grow faster comapred to single-agent or single-author scenarios, only well-funded actors may afford the energy
budget, deepening the resource gap already highlighted for modern NLP
pipelines \cite{Strubell2019Energy}.  Sustaining the benefits of
discursive verification therefore demands not merely algorithmic
innovation but also governance frameworks and infrastructural subsidies
that keep the playing field environmentally and economically fair.

Discursive networks promise robust cross-verification, yet their
perpetual negotiation of ``truth" can erode epistemic diversity.
When many agents iteratively revise one another, the process tends to
pull answers toward a central consensus, suppressing minority
explanations in favour of the statistically safest wording.
Large-scale language models already exhibit this homogenising bias,
along with well-documented tendencies to replicate and even amplify the
social prejudices embedded in their training data
\cite{Bender2021Parrots}.  A network of such agents therefore risks
hard-coding bias under the reassuring veneer of multi-agent agreement.

At the same time, cross-agent review
typically demands full visibility of prompts and intermediate
reasoning, thereby increasing the attack surface for privacy breaches.
Membership-inference studies demonstrate how seemingly benign queries
can reveal whether sensitive records were present in a training set
\cite{Shokri2017Membership}, suggesting that discursive networks must
treat every inter-agent channel as a potentially hamful vector for leakage.

Finally, adversarial robustness and distributive justice pose
intertwined challenges.  A single compromised agent can inject
fashioned ``triggers" that, once propagated through mutual validation,
shift the entire network toward a malicious conclusion
\cite{Wallace2019Triggers}.  

\subsection{Conclusions and Outlook}
\label{sec:conclusion}

This manuscript has traced a broad arc, from theoretical grounding to
practical tooling, around a single organising idea: \emph{discursive
networks}.  By recognising that every LLM is
both a generator and a consumer of discourse, we cast its interactions as
edges in a network whose universal structure can be exploited for
robust error control.   
Building on this abstraction we 
introduced \emph{Flaws-Of-Others}~(FOO), a reconfigurable
agent-based algorithm packaged with user-friendly tools that assist the
production, verification, and revision of scientific knowledge.  

Many ethical challenges in discursive networks call  for detailed, tamper-proof logs of interactions. This is essential for  accountability, reproducibility, and auditability in digital systems. When interactions involve LLMs, scientific collaborations, or complex data workflows, the capacity to verify the integrity of recorded exchanges becomes a prerequisite for trust. One approach to achieving this is through the use of blockchain technology, which can provide decentralized, cryptographically secure records that are resistant to unauthorized modification. Each entry in a blockchain-based log is linked to the previous one through cryptographic hashes, ensuring that any tampering with earlier data invalidates the entire subsequent chain. This structure allows, in principle, interaction logs to be both transparent and verifiable without central oversight. Furthermore, incorporating time-stamping and access control into such systems ensures that each interaction is both temporally fixed and attributable. These properties make blockchain a viable framework for securing interaction logs in research, legal compliance, and automated decision-making contexts.


This study deliberately replaces the fashionable term
\emph{hallucination} with the broader concept of
\emph{invalidation}.  Whereas ``hallucination" suggests a purely
accidental slip in the model's internal perception, the data reveal a
richer spectrum of failure modes that includes strategic prompt manipulation, chain-of-thought drift and the simple inheritance of errors from flawed training corpora.  All of these mechanisms manifest in the observable metric that matters to users, the production of false statements, so the single hazard rate \(\lambda\) is most naturally interpreted as an \emph{invalidation rate}.  The shift in vocabulary is therefore more than semantics: it aligns theoretical parameters with the phenomena actually counted in benchmarks.

A discursive network that relies exclusively on its own
self-correction routines will remain \emph{invalidation-dominant} as soon
as the fabrication hazard exceeds the internal repair rate
(\(\lambda>q\)).  In that regime the long-run share of false statements
is bounded below by one half, regardless of implementation details or
domain.  The formalism developed here reveals why: fabrication and
self-repair enter the steady-state ratio
\(\pi_f^{\text{single}}=(p+\lambda)/(p+\lambda+q)\) additively, leaving
no structural mechanism for the network to ``outrun" its own
invalidations.

The picture changes once independent verification channels are added.
Coupling the generator to \(n-1\) external agents augments every false
statement with an additional repair hazard \(d\) per agent, yielding the
effective rate \(q_{\text{eff}}(n)=q+(n-1)d\).
If the composite system satisfies
\(\lambda/d < p/(p+q)\) it flips into the
\emph{truth-dominant} regime, in which the falsehood share decreases
monotonically with \(n\) and approaches zero in the limit of infinite
cross-checking capacity.  The transition threshold depends only on the
ratio \(\lambda/d\), providing a clean design criterion that is
independent of any particular benchmark or simulation.

Proposition~\ref{prop:n_agents} turns that qualitative
criterion into a quantitative planning tool.  It gives a closed-form
bound
\[
  n_{\min}
  =
  \Bigl\lceil
    1+\frac{(p+\lambda)\bigl(\tfrac1\varepsilon-1\bigr)-q}{d}
  \Bigr\rceil
\]
for the smallest number of mutually detecting agents required to keep
the long-run falsehood share below a user-specified tolerance
\(\varepsilon\).  Because \(n_{\min}\) grows only \emph{linearly} in
\((p+\lambda)/d\), even ambitious error targets (e.g.\ \(\varepsilon=0.01\))
translate into tractable network sizes.  In practice, engineers can
estimate \(p\), \(q\), and \(\lambda\) from established benchmarks,
select a viable cross-detection mechanism to determine \(d\), and then
read off \(n_{\min}\) directly from the formula.  The discursive-network
formalization thus provides not just a descriptive model of information
dynamics but a concrete apparatus for right-sizing the verification
infrastructure needed to achieve prescribed levels of factual
reliability.

The theoretical analysis confirms that our mathematical framework produces stable, interpretable dynamics across parameter ranges consistent with published LLM studies. The models successfully capture qualitative regime transitions (invalidation-dominant vs. truth-dominant) and provide quantitative predictions for multi-agent system design. 

Empirical validation remains an important direction for future work, requiring controlled experiments specifically designed to measure the theoretical parameters ($p$, $q$, $\lambda$, $d$) under realistic conditions.

Although the mathematics treats \(d\) as a single scalar, the concept
encompasses several concrete engineering choices.  Retrieval-augmented
generation raises \(d\) by surrounding the model with authoritative
passages that expose contradictions.  Model-ensemble adjudication raises
it further by combining the diverse priors of independently trained
models; uncorrelated errors rarely agree, so the aggregate chance that a
falsehood slips through diminishes rapidly.  Even higher values of
\(d\) become attainable when a human editor is placed in the loop,
although latency and cost then become limiting factors.  Our hazard
model quantifies these trade-offs: for any target false-statement
tolerance one can compute a required detection rate, and thus budget the
amount of human or automated scrutiny that must be applied.

Just as people are often quicker to spot another person's mistakes than their
own, language-model systems can sometimes be more useful as critics of a peer's
text than as unchecked authors.  The mathematical claim, however, is conditional
rather than universal.  Cross-model review helps when the reviewer has access to
a validity signal that the authoring pass did not use: external evidence,
retrieval, tools, formal constraints, process supervision, independent prompts,
or genuinely diverse model families.  It can fail when all agents share the same
blind spot.  The FOO architecture therefore treats critique not as a magical
property of LLMs but as an engineered verification channel whose sensitivity,
false-positive rate, repair probability, and common-mode dependence must be
estimated.

The significance of these findings extends beyond artificial systems.
Invalidation dynamics of the same mathematical form could govern human
conversation, peer review and social-media fact-checking.  Recognising
this shared structure invites a unified research agenda that links
network science, cognitive psychology and algorithmic governance.  Our
future work will explore heterogeneous hazards at the level of
individual actors, time-varying detection capacities that respond to
workload, and live A/B tests that recover hazard estimates directly from
production chat traffic.  A deeper ethical analysis will also be
required, because the push toward smaller \(\pi_{F}\) competes with
privacy constraints, energy budgets and the carbon footprint of
large-scale verification.

\paragraph{We have entered a new cultural mode. It affects scientific production.} 
We have crossed a cultural threshold: language is no longer crafted solely by human hands and minds, but is continuously co-composed with machines.  Large language-model systems do not ``assist'' writing; they repurpose it.  Authorship dissolves into a live dialogue between human intention and algorithmic completion, with sentences generated, revised, and re-externalized in the same breath.  Because the practice itself has changed, the evaluative yardsticks built for solitary, page-bound prose (originality scores, citation counts, style rubrics, etc.) no longer capture what is happening on the screen.  This cultural shift brings novel risks and opportunities, demanding updated methods, training, expectations, and metrics.

Looking forward, the research programme must widen from
token-level truthfulness to \textbf{medium-level dynamics}.  Three paths
stand out:

\begin{enumerate}
    \item \textbf{Cultural assimilation.}  Our most pressing need is  an ethical framework to harmonize societal values with this new reality.  We need conventions,
          interfaces and pedagogies that make continuous, model-mediated
          composition legible, trustworthy, and fair. 
    \item \textbf{Metric redesign.}  Benchmarks rooted in solitary
          authorship and static text can no longer capture quality in a
          live, co-creative medium.  New metrics should score how a
          statement evolves under iterative detection and repair, not
          just its instantaneous truth value.
    \item \textbf{Governance of agency.}  When the medium itself
          ``acts on the message," responsibility diffuses across
          designers, deployers and end-users.  Future hazard models
          must therefore integrate economic incentives, interface
          affordances and policy constraints alongside the technical
          parameters \(p,q,\lambda,d\).
\end{enumerate}

Seen through this lens, the ideas discussed in this manuscript form a
prototype for broader cultural discussions that will emerge as society
learns, \emph{once again}, to write in a fundamentally new medium.

\section*{Acknowledgements}
This work was supported in part by the National Institute of General
Medical Sciences of the National Institutes of Health under Award
No.\ \href{https://reporter.nih.gov/project-details/10723223#history}{1R25GM151182} and by the National Science Foundation under Award 
No.\ \href{https://www.nsf.gov/awardsearch/showAward?AWD_ID=2518973&HistoricalAwards=false}{2518973}.  The content is solely the responsibility of the authors
and does not necessarily represent the official views of the NIH, NIGMS,
or NSF.

In an instance of self-reference, this manuscript was developed using the methods it describes, through iterative cycles of recursive software development that refined the underlying technology. The records of interactions with multiple LLM agents are recorded only for late-stage manuscript refinement. Nevertheless, the following statement accurately represents the process employed here and provides a template for future work:

\begin{quote}
This manuscript and its supplementary materials were produced using methods detailed in \cite{gutierrez2025flaws}. The author supplied the core concepts, the logical framework, and foundational technical content, while large language-model assistance was utilized for ideation, verification, text drafting and revision, and software implementation. Responsibility for all claims made herein rests solely with the author. Existing logs of interactions with LLM agents are included as an appendix within the supplementary materials.
\end{quote}

\printbibliography

\newpage
\appendix
\section{Blockchain Implementation Details}\label{appendix:blockchain-implementation}. 

\subsection{Anchoring options for the FOO audit log}
\label{app:anchoring}

A signed hash-linked log is tamper-evident only relative to a trusted copy of
the log head.  There are three standard ways to create such a reference point.
A trusted time-stamping authority can issue an RFC~3161 time-stamp token
asserting that the current head hash existed at a specified time
\parencite{rfc3161}.  A public-chain anchoring service can periodically commit a
Merkle root containing FOO log heads to a public ledger; this outsources
append-only evidence to the ledger's consensus mechanism.  Finally, an
organization can maintain the log head through a Byzantine-fault-tolerant
replication protocol such as PBFT, thereby tolerating a bounded number of faulty
or malicious replicas \parencite{castro1999pbft}.  Without one of these
anchoring mechanisms, a local hash chain should be described as tamper-evident
under comparison to a stored head, not as tamper-proof.

\paragraph{Implementation note.}
We implement the loop in \texttt{Python~3.11} with \texttt{asyncio} concurrency; each agent call is an HTTP request to a hosted LLM endpoint (e.g.\ OpenAI, Anthropic).   Neither the number of agents nor their prompts are fixed: both are read at run time from a simple JSON configuration, so the same engine can mediate anything from a two-model A/B test to a dozen specialized critics.  Unless stated otherwise, the experimental pool comprises a single \emph{harmonizer} (the most capable model available) and, for \emph{each} back-end engine under test, two specialist agents sampled at temperatures~0.1 and~0.9.  The protocol executes one broadcast round plus a minimum of three consensus rounds, so every query triggers
\[
\text{cost} = 4 \times |A| \quad \text{API calls},
\]
where \(|A|\) is the total number of agents (harmonizer + specialists).
For example, with one engine (\(|A|=3\)) the loop issues
\(4\times3 = 12\) calls; adding engines or additional specialist roles
scales the cost linearly.  
All configuration files and source code are publicly available at
\url{https://github.com/biomathematicus/foo}.

\paragraph{Blockchain motivation.} 
The reliability of discursive networks depends critically on the integrity of recorded interactions between human and artificial agents. When scientific conclusions emerge from iterative exchanges among multiple LLMs and human reviewers, the provenance and authenticity of each contribution becomes essential for both reproducibility and accountability. Traditional logging systems are vulnerable to post-hoc modification, making it difficult to distinguish genuine collaborative refinement from retrospective tampering or fabricated authorship claims.


We address this challenge through a blockchain-based integrity system that creates tamper-evident records of all agent interactions. Each message exchange in the discursive network generates a cryptographic block containing the agent identity, message content, timestamp, and hash-linked reference to the previous interaction. The system employs SHA-256 hashing with a global salt shared across all agents to ensure consistency while preventing individual agents from being identified through hash analysis alone.

\begin{definition}[Conversation Blockchain]
For a discursive network defined by the structure $N = (S,P,I,C,A,B,U,G)$, the conversation blockchain $\mathcal{B}$ is a sequence of cryptographically linked blocks $\{b_0, b_1, \ldots, b_n\}$ that creates tamper-evident records of all agent interactions, where $\sigma$ is a global salt and $||$ denotes concatenation.
\end{definition}

\begin{algorithm}[htb]
\caption{Blockchain record creation for discursive networks}
\label{algo:blockchain}
\begin{algorithmic}[1]
\Require message content $m_i$, sender $a_j$, receiver $a_k$, global salt $\sigma$
\Ensure new blockchain block $b_i$
\State $t_i \gets$ current timestamp
\State $h_c \gets H(m_i || t_i || \sigma)$ \Comment{Content hash}
\If{$i = 0$} \Comment{Genesis block}
    \State $h_b \gets H(h_c || \text{``GENESIS''} || \sigma)$
\Else
    \State $h_b \gets H(h_c || h_b(b_{i-1}) || \sigma)$ \Comment{Chain hash}
\EndIf
\State $b_i \gets \{m_i, t_i, h_c, h_b, i, \text{verified}\}$
\State Append $b_i$ to blockchain $\mathcal{B}$
\State \Return $b_i$
\end{algorithmic}
\end{algorithm}

\begin{algorithm}[htb]
\caption{Blockchain integrity verification}
\label{algo:blockchain-verify}
\begin{algorithmic}[1]
\Require blockchain $\mathcal{B} = \{b_0, b_1, \ldots, b_n\}$, global salt $\sigma$
\Ensure integrity status (valid/tampered)
\For{$i = 0$ to $n$}
    \State $h_c^* \gets H(m_i || t_i || \sigma)$
    \If{$h_c^* \neq h_c(b_i)$}
        \State \Return ``TAMPERED: Content hash mismatch at block $i$''
    \EndIf
    \If{$i > 0$}
        \State $h_b^* \gets H(h_c(b_i) || h_b(b_{i-1}) || \sigma)$
        \If{$h_b^* \neq h_b(b_i)$}
            \State \Return ``TAMPERED: Chain hash mismatch at block $i$''
        \EndIf
    \EndIf
\EndFor
\State \Return ``VERIFIED: Blockchain integrity intact''
\end{algorithmic}
\end{algorithm}

The cryptographic hash function $H(\cdot)$ can be instantiated with various secure hash algorithms including SHA-256, SHA-3 (Keccak), BLAKE2, or BLAKE3, each offering different performance and security trade-offs. SHA-256 remains a robust choice for blockchain applications due to its widespread adoption and proven security properties, while newer algorithms like BLAKE3 offer superior performance for high-throughput scenarios.

The blockchain protocol ensures that any modification to historical interactions invalidates the cryptographic chain, producing detectable integrity violations. When agents load previous conversations, the system verifies the complete hash chain and displays prominent warnings if tampering is detected: ``LOG TAMPERED. TRUST HAS BEEN BREACHED. BLOCKCHAIN FAILS.'' This mechanism makes post-hoc fabrication of contributions computationally infeasible while preserving the ability to legitimately edit conversations by rebuilding the chain from the point of modification onward.

The implementation maintains separate blockchains for each agent while using a shared salt stored in the system configuration to ensure hash consistency across sessions. Genesis blocks are initialized with fixed timestamps to prevent hash divergence during system restarts. The protocol automatically migrates existing conversation logs to blockchain format, enabling backward compatibility while establishing integrity verification for all new interactions.

This cryptographic foundation serves two complementary functions in discursive networks. First, it provides technical infrastructure for reproducible research by creating verifiable logs of how scientific conclusions evolved through agent interactions. Second, it establishes an ethical framework for attribution by making it computationally expensive to falsify contributions after the fact. The blockchain thus transforms the question of authorship from a matter of trust to one of cryptographic verification, supporting the broader goal of maintaining accountability in collaborative human-AI knowledge production.

\end{document}